\documentclass[final,1p,times]{elsarticle}

\usepackage{amssymb}
\usepackage{amsthm}

\usepackage{graphicx}
\graphicspath{{figures/}}

\usepackage{array,bm}
\usepackage{xcolor}
\usepackage{float}
\usepackage{subcaption}
\usepackage{url}
\usepackage{siunitx}
\usepackage{verbatim}
\usepackage[justification=centering]{caption}
\usepackage{gensymb}
\usepackage{hyperref}
\usepackage{xparse}
\usepackage{siunitx}
\usepackage{amsmath}
\usepackage[export]{adjustbox}

\makeatletter
\def\ps@pprintTitle{%
  \let\@oddhead\@empty
  \let\@evenhead\@empty
  \def\@oddfoot{\reset@font\hfil\thepage\hfil}
  \let\@evenfoot\@oddfoot
}
\makeatother

\begin{document}

\begin{frontmatter}



\title{Validation of a photogrammetric approach for the objective study of early bowed instruments}


\author[UCL,ICTEAM]{Ph. Beghin}
\ead{philemon.beghin@uclouvain.be} 
\author[UCL,INCAL,MIM]{A.-E. Ceulemans}
\ead{anne-emmanuelle.ceulemans@uclouvain.be} 
\author[UCL,IMMC]{P. Fisette}
\ead{paul.fisette@uclouvain.be} 
\author[UCL,ICTEAM]{F. Glineur}
\ead{francois.glineur@uclouvain.be} 

\affiliation[UCL]{organization={UCLouvain},
            city = {Louvain-la-Neuve},
            country={Belgium}}
            
\affiliation[ICTEAM]{organization={Institute of Information and Communication Technologies, Electronics and Applied Mathematics (ICTEAM)}}

\affiliation[IMMC]{organization={Institute of Mechanics, Materials and Civil Engineering (IMMC)}}

\affiliation[INCAL]{organization={Institute for the Study of Civilisations, Arts and Letters (INCAL)}}

\affiliation[MIM]{organization={Musical Instruments Museum (MIM)},
             city={Brussels},
             country={Belgium}}

\begin{abstract}

\noindent Some early violins have been reduced during their history to fit imposed morphological standards, while more recent ones have been built directly to these standards. We propose an objective photogrammetric approach to differentiate between a reduced and an unreduced instrument, whereby a three-dimensional mesh is studied geometrically by examining 2D slices. Our contribution is twofold. First, we validate the quality of the photogrammetric mesh through a comparison with reference images obtained by medical imaging, and conclude that a sub-millimetre accuracy is achieved. Then, we show how quantitative and qualitative features such as contour lines, channel of minima and a measure of asymmetry between the upper and lower surfaces of a violin can be automatically extracted from the validated photogrammetric meshes, allowing to successfully highlight differences between instruments.
\end{abstract}




\end{frontmatter}



\section{Historical context and motivation}
\begin{figure}[H]
  \centering
  \begin{subfigure}{0.27\linewidth}
    \centering
    \includegraphics[scale = 0.24]{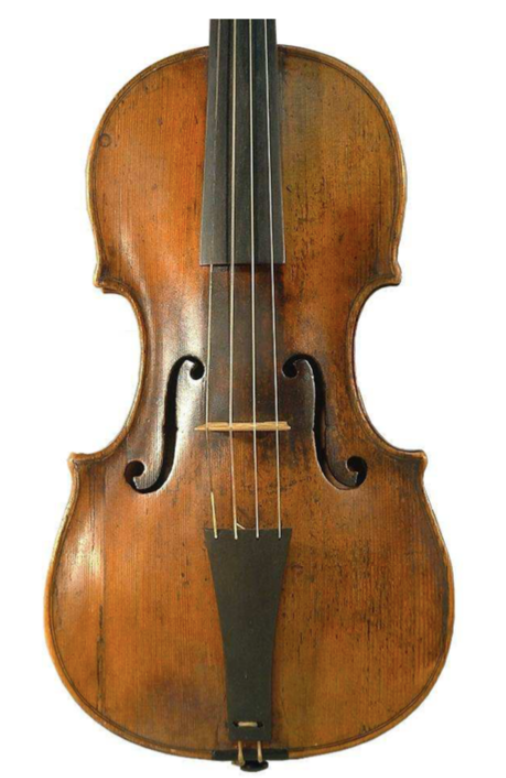}
  \end{subfigure}%
  \hspace{-1cm}
  \begin{subfigure}{0.24\linewidth}
    \centering
    \includegraphics[scale = 0.25]{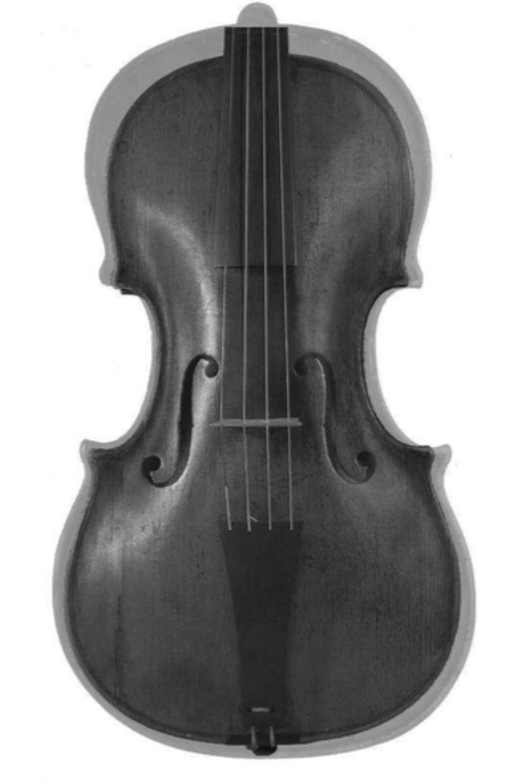}
  \end{subfigure}%
  \hspace{1cm}
  \begin{subfigure}{0.27\linewidth}
    \centering
    \includegraphics[scale = 0.088]{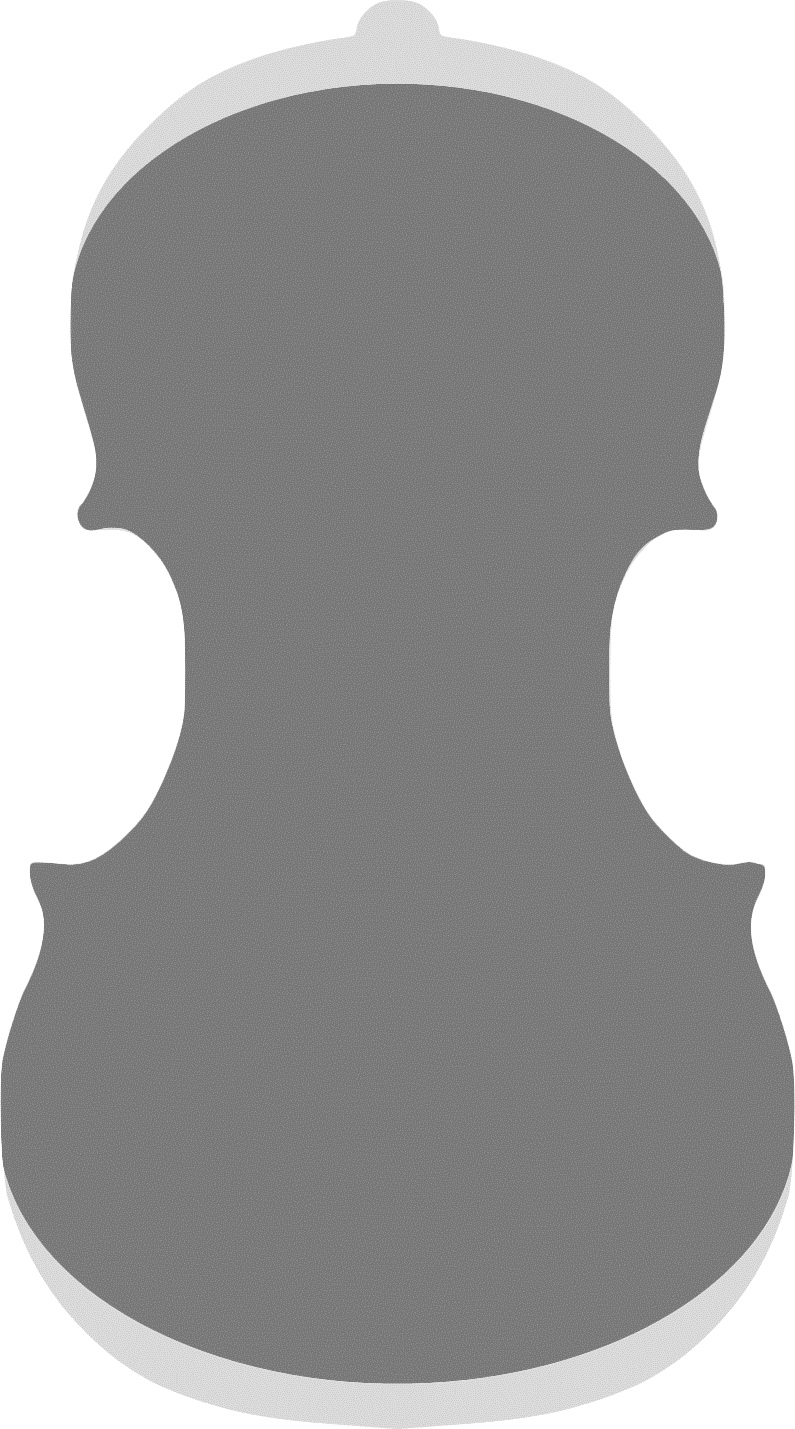}
  \end{subfigure}
  \hspace{-1cm}
  \begin{subfigure}{0.24\linewidth}
    \centering
    \includegraphics[scale = 0.09]{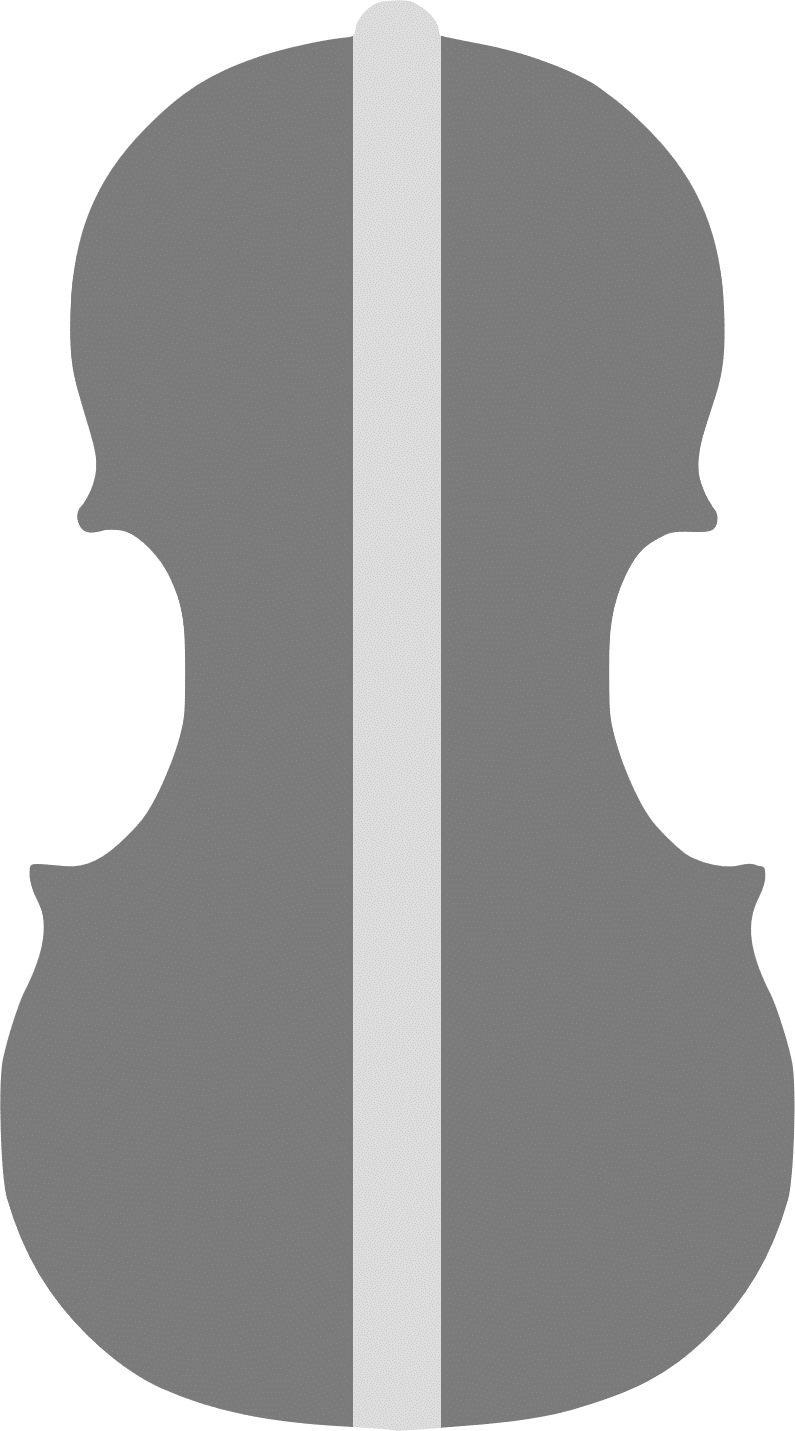}
  \end{subfigure}  
  \caption{Reduced violin vs. its estimated original dimensions (left) \cite{moens2015voix} \\
  Reduction of the height of the sound box vs. reduction of the width (right).}
  \label{fig:Moens}
\end{figure}

\noindent The morphology of today's violin differs greatly from that of the  instruments built between the late 16th and the mid-18th century. After 1750, in order to meet the standards imposed by famous orchestras and conservatories, many early violins have been reduced. Figure \ref{fig:Moens} shows on the left a reduced violin from the first half of the 18th century and an estimation of its original dimensions \cite{moens2015voix}. 
We illustrate two types of reduction on the right: re-cutting the top and bottom parts of the violin body (also called the sound box), and removing a slice of wood along the axis of the instrument to reduce its width. 
As historical testimonies about this process are imprecise, a common issue for today's musicologists, organologists and luthiers is to determine whether an early violin has been reduced and, if so, to quantify the alterations it has undergone. This problem has been little studied but is nevertheless important because it changes the image of pre-1750 music. A detailed historical account of this issue can be found in \cite{GPS}. It is therefore desirable to evaluate the violin geometry in a completely objective way, which is the problem we address. \\

\noindent Our aim is to detect differences between reduced and unreduced violins on the basis of 3D models representing the instruments, since the violins themselves are the best witnesses of their morphological evolution (rather than written sources, for instance). Two particular cases of reduced violins attributed to Andrea Amati bear a painted heraldic shield. A complete study \cite{radepont2020revealing} of the modified coat of arms (using notably X-ray fluorescence spectroscopy and historical knowledge) allowed to identify precisely how the instruments were reduced. Unfortunately, this approach is rather unique and cannot be generalised, as most violins are devoid of pictural ornamentation.

\setlength{\belowcaptionskip}{-10pt}

\noindent To the best of our knowledge, no additional work on the quantification of reduction of early violins has ever been carried out. We can however mention studies related to the violin, such as the 2D classification \cite{chitwood2014imitation} describing the morphological evolution of the violin body over 400 years of history (depending on time, luthier style, geographic area, etc.). This work, performed on top view pictures of more than 9000 instruments, aimed to isolate and study the contours of the violins. These contours were represented with elliptical Fourier descriptors and then classified using Principal Component Analysis (PCA) and Linear Discriminants. The study has finally shown that violin shapes tend to cluster into four major groups based on factors such as dimensions, curvatures and bout placement. An extension \cite{peron2018pattern} has been proposed to a larger and better available database, namely the Musical Instrument Museums Online\footnote{\url{https://mimo-international.com/MIMO/}} (MIMO), which offers information on numerous instruments held in public museums. From the violin images, the authors derived a set of measurements that reflect relevant geometric features of the instruments. The application of PCA uncovered similarities between violin makers and their respective copyists, as well as among luthiers belonging to the same family lineage, in the context of a historical narrative. \\

\noindent Other researchers have used deep learning and convolutional neural networks (CNN) for stylistic recognition of historical violins \cite{dondi2021stylistic}. The CNNs had to automatically determine whether an instrument was made by Antonio Stradivarius or not (binary classification). Photos were given as input, focusing on either the violin body, the head or both. Once again, an exclusively 2D approach was implemented whereas our goal is to consider a 3D model. \\

\noindent None of the aforementioned works are concerned with violin reduction. However, in contrast to the literature on instrument reduction, several studies have been performed on the reconstruction of 3D models of violins (either for actual modelling or for vibro-acoustic applications). Techniques used include laser scanning \cite{dondi2016measuring, dondi20173d}, medical X-ray computed tomography (CT) scans \cite{lothairecharacterization, kirsch2015x, stanciu2021x, pyrkosz2011converting, pyrkosz2013comparative, pyrkosz2014coupled}, neutron imaging \cite{kirsch2015x}, high resolution industrial CT scans ($\mu$-CT scans) \cite{plath2017post, plath20193d, marschke2018modeling, marschke2020approach, fuchs2018musices}, UV fluorescence with the use of a Kinect device \cite{dondi2018multimodal} and finally, photogrammetry \cite{pinto2008photogrammetric}. Photogrammetry consists of digitally recreating a 3D object based on 2D photographs. It is this last technique, the most accessible, that we have focused on. \\

\noindent The advantages of photogrammetry are that it is non-invasive to instruments and that it is a mobile technique \cite{plath20193d}. A study in which measurements were made on a photogrammetric 3D model of a violin and then compared to a synthetic version of that violin showed that the reconstructed surface matched the model with an average error of a few hundredths of mm \cite{pinto2008photogrammetric}, encouraging confidence in photogrammetry. Furthermore, this technique has already made its mark in several other areas. In a medical context, it offers an alternative to scanners for patients who are too sensitive to radiation \cite{liu2019validation, ho2017comparing}. This use was validated by comparison to CT scans using statistical tools such as Bland-Alman graphs \cite{donato2020photogrammetry} or the Student $t$ test \cite{ho2017comparing}. We aim to validate it here with CT scans by means of geometric properties instead of statistical ones. \\

\noindent Several studies have recreated 3D models of violins, but none have analysed them or examined their morphology. In the future, we would also like to be able to predict how the instruments were reduced, hence the importance of having accurate models. \\

\noindent In this paper, we study two violins\footnote{Strictly speaking, both of them are violas (the intermediate violin size), but we will use the generic name of `violin' throughout this article.}, one of which is strongly suspected to be reduced. Our main tools are 3D photogrammetric and CT scans meshes whose acquisition is described in Section \ref{Acquisition}. We validate the use of the photogrammetric meshes by estimating how accurate they are with respect to CT meshes in Section \ref{Validation}. Finally, in Section \ref{Geometric analysis}, we use the validated photogrammetric meshes to highlight the contour lines of the violins, their minima channel and the asymmetry between the upper and lower surfaces of their body, allowing us to illustrate differences between a reduced and an unreduced instrument. \\

\section{Mesh acquisition}
\label{Acquisition}

\noindent Both studied instruments will be referenced by their luthier's name: \textbf{Hofmans}\footnote{Hofmans Matthys IV, inv. no 2846, Antwerp, before 1679 (Musical Instruments Museum Brussels).} (which is believed to be reduced) and \textbf{Cuypers}\footnote{Cuypers Johannes Th., inv. no 2833, The Hague, 1761 (Musical Instruments Museum Brussels).} (which is not). As the necks have been replaced over time \cite{stowell1990violin}, we focus exclusively on the upper and lower surfaces of their body, respectively called the `sound board' (not to be confused with the sound box) and the `back'. In this section, we first describe the two methods with which we acquire our meshes, and then show how to isolate the sound board or back of the instruments for fair comparisons between representations. 

\subsection{Photogrammetric mesh}
\label{Photogrammetric mesh}
\begin{figure}[H]
\centering
\includegraphics[scale = 0.0255]{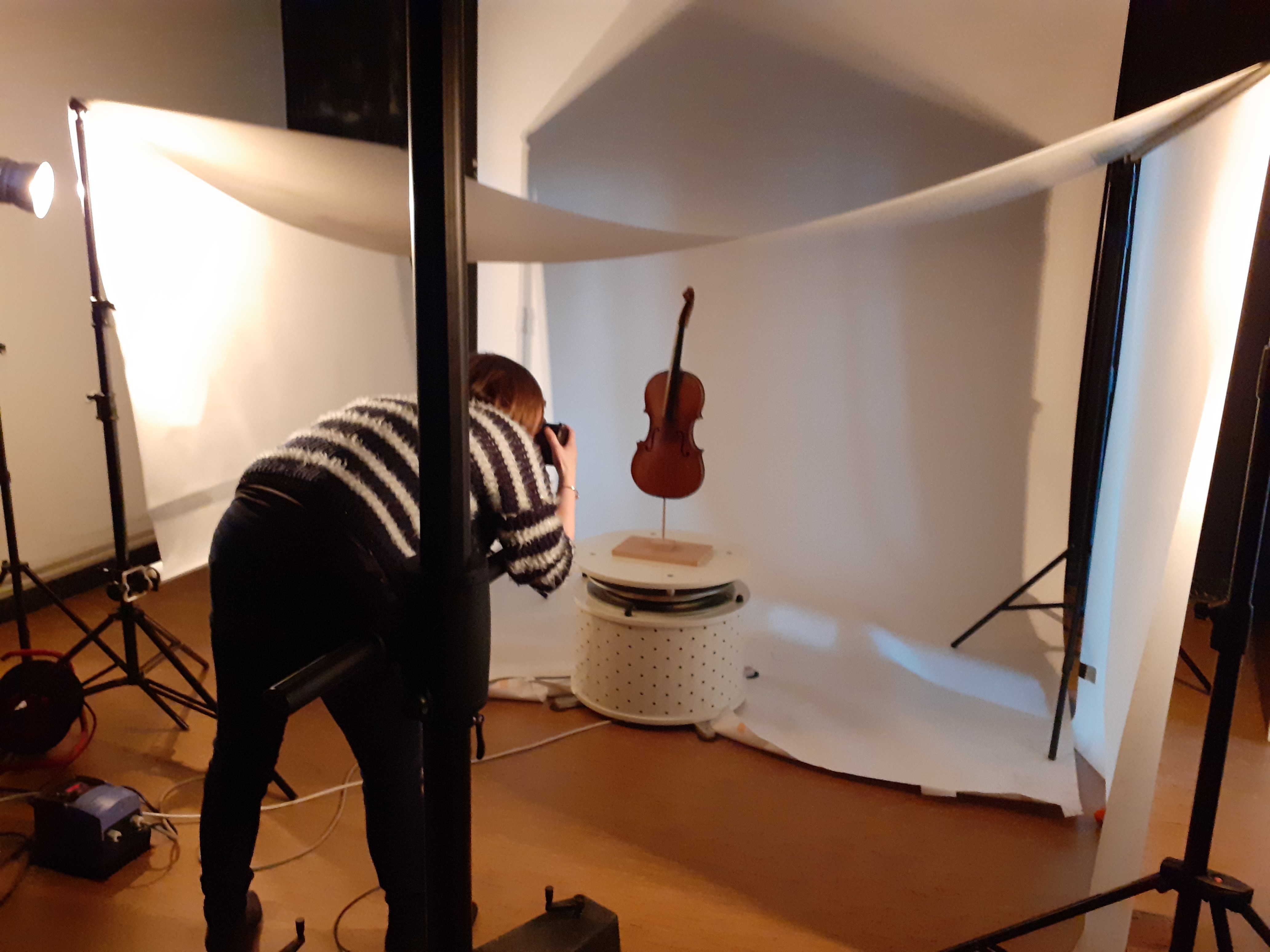}
\includegraphics[scale = 0.06]{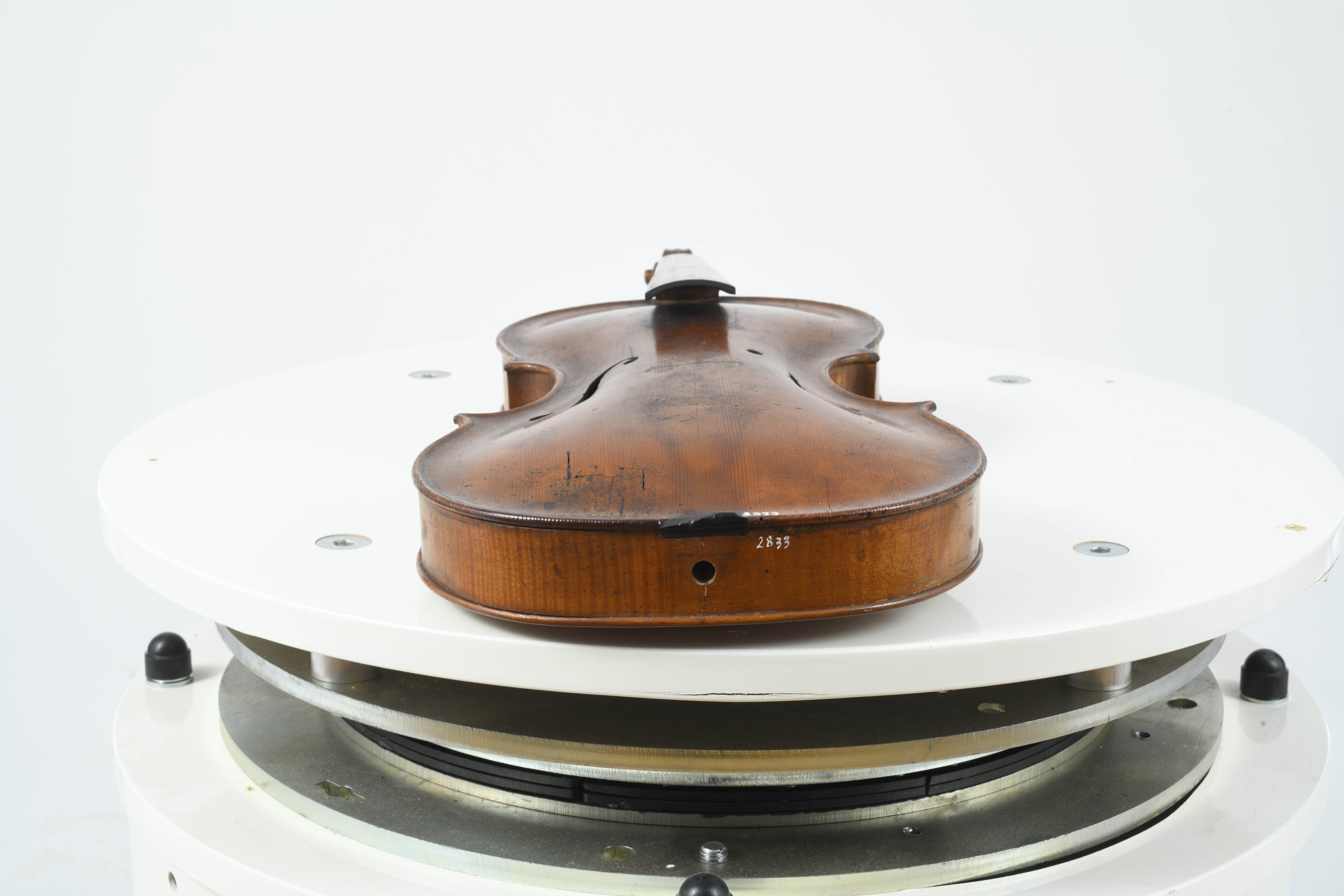}
\includegraphics[scale = 0.06]{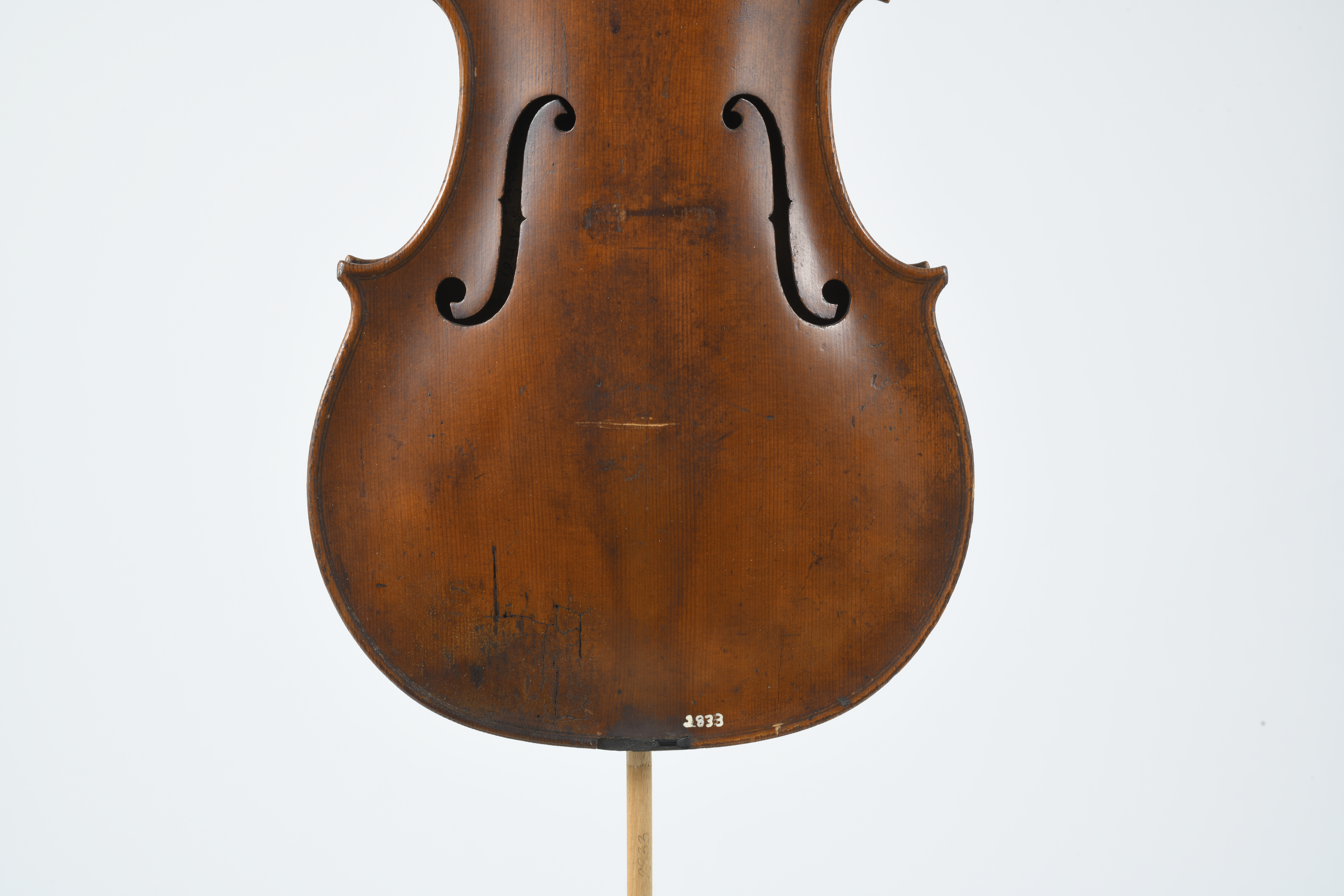}
\caption{Setup and photographed violin (left: light tent, centre: laid down, right: upright).}
\label{fig:Setup}
\end{figure}

\noindent We have benefited from the valuable help of Iona Thys, photographer at the Royal Museums of Art and History (Brussels), to create the photogrammetric models. About 160 photos for each instrument were taken by a Nikon D850 camera with a \SI{60}{\mm}  focal lens. The two violins were placed on an automatic turntable, rotated through $\ang{360}$ and photographed every $\ang{10}$ from three different perspectives (heights). Each picture contains $8256 \times 5504$ pixels ($\approx$ 50--60  $\mu m$ per pixel) and is about $20$ MB. The software that creates the 3D model must receive enough information, and especially overlapping pictures. The main challenges encountered during our photogrammetry campaign are related to lighting, since photogrammetry cannot reconstruct varnished, reflecting or transparent surfaces. Hence, a light tent was used to provide an indirect soft lighting and avoid strong reflections due to the violin varnish (see Figure \ref{fig:Setup} (left)). Both instruments were photographed lying down on the turntable and upright on a wooden stick (still at different heights), as can be seen in Figure \ref{fig:Setup} (centre, right). 

\noindent Once all the pictures were taken, their background was eliminated with Adobe Photoshop\footnote{\url{https://www.photoshop.com/en}}. This procedure aims to delineate each instrument with a mask, which is helpful in the meshing process to better detect the key points of the violin and its contour. Each masked photo is also double-checked and adjusted in case the automated masking procedure has failed and retained some artefacts. The mask of each violin picture and the original images are then sent to the photogrammetry software Agisoft Metashape\footnote{\url{https://www.agisoft.com}} to create the meshes. When creating the model, Metashape takes into account the relative measurements of the violin, but the software is unable to calculate the actual dimensions of the instruments. Thanks to the RadiAnt DICOM Viewer software\footnote{\url{https://www.radiantviewer.com/}}, we can measure distances on the CT scans. By averaging a few typical distances, we scale up the photogrammetric mesh to make it correspond to the actual dimensions of the violin. We will explain in Section \ref{Registration between photogrammetric and CT representations} how we have corrected this manual scaling. Eventually, the sound board and the back are separated (more on this in Section \ref{Contour isolation}) and the sound holes are delineated and removed manually. The resulting sound board and back meshes contain about 400k--500k vertices.

\subsection{CT scan mesh}
\label{CT scan mesh}
\noindent Both violins were scanned at the University Hospital Saint-Luc (UCLouvain, Brussels-Woluwe), which produced $512 \times 512$ pixels slices with an overlap rate of $50\%$ (around $2300$ slices with \SI{0.67}{\mm} thickness for Hofmans and $1600$ slices with \SI{0.9}{\mm} thickness for Cuypers). The medical images were then converted into meshes using the ITK-SNAP\footnote{\url{http://www.itksnap.org/pmwiki/pmwiki.php}} software based on the contour segmentation algorithm detailed in \cite{py06nimg}. As with the photogrammetric meshes, the sound board and the back were separated. In addition, the part of the mesh corresponding to inner walls was removed manually. Indeed, unlike photogrammetry which only acquires the outer surface of an object, CT scans detect their inner surfaces as well. The resulting sound board and back meshes contain about 330k–430k
vertices. \\

\noindent The use of CT scans for the two instruments we study here was available thanks to the work conducted in \cite{lothairecharacterization}. Unfortunately, despite the good accuracy they provide, the use of medical scanners is somehow restrictive. First, all the instruments brought from the museum to the hospital are historical artefacts which need to be insured. Then, the number of instruments that can be scanned is limited and the scanners themselves must remain available in case of medical emergency. For obvious ethical reasons, it is difficult to find a time slot for this type of research when a patient’s health may be at stake. Moreover, two technologists must be present. The first one adapts the scanner settings to the density and material of the violin wood while the second handles the scanner itself. Finally, some instruments carry pathogens and cannot be scanned at all. As we plan to extend our research to a corpus of about forty more instruments, all those reasons led us to consider a simpler way to proceed and motivated us to focus on photogrammetry, which we will validate with the CT scan information we already own.

\subsection{Contour isolation}
\label{Contour isolation}

\noindent Before validating our photogrammetric meshes by comparing them to the CT scan meshes, we need to make sure that we are dealing with similar digital representations of the objects. We are mainly interested in the sound board and back of the violin, and have developed an automatic method to `delineate' these two surfaces from a complete instrument mesh, which is not a trivial problem. \\

\begin{figure}[H]
    \centering
    \begin{subfigure}{\textwidth}
     \centering
    \includegraphics[width = 11 cm]{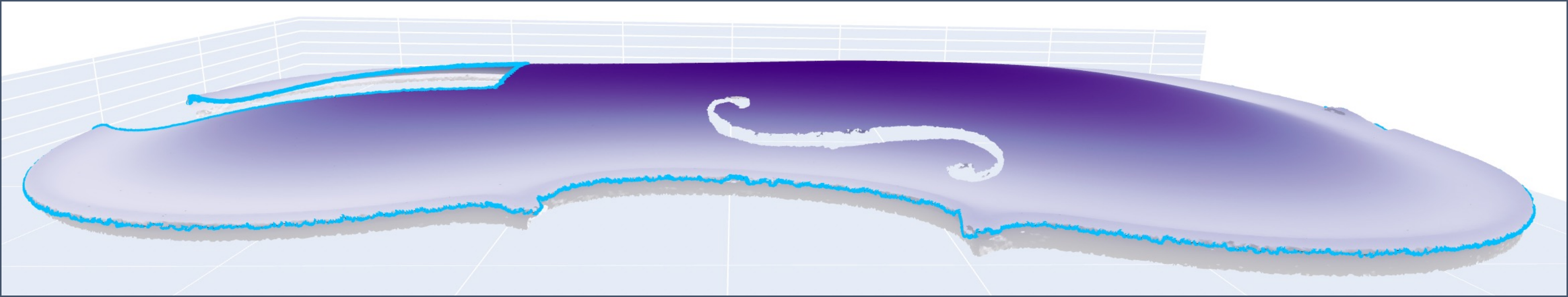}
    \end{subfigure}
    \begin{subfigure}{\textwidth}
    \vspace{0.25 cm}
     \centering
    \includegraphics[width = 11 cm]{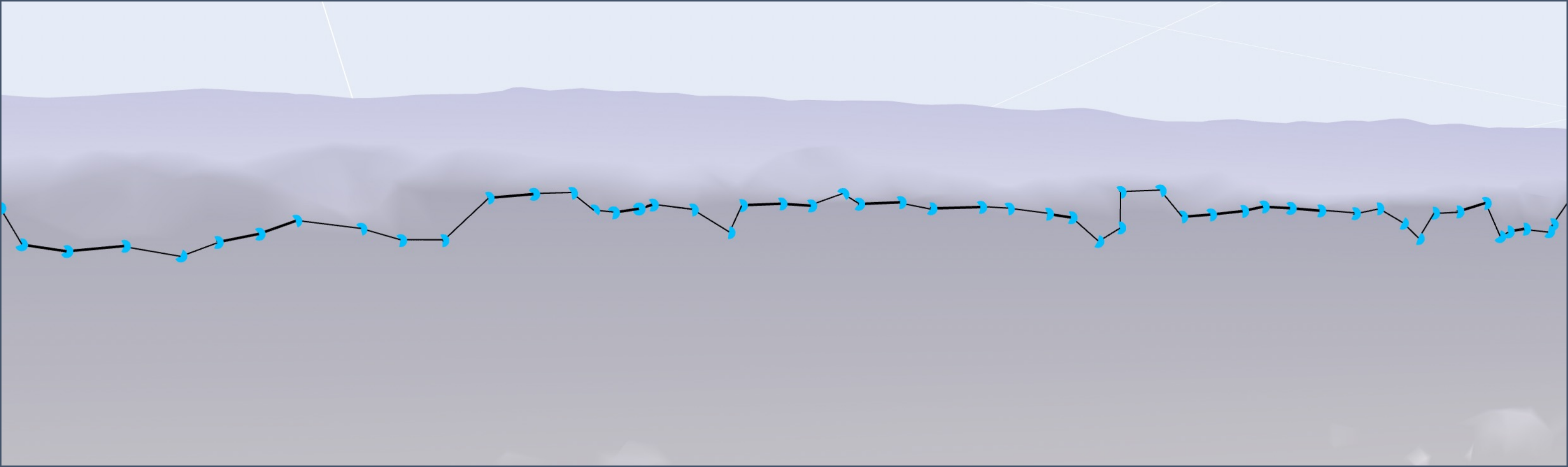}
    \end{subfigure}
    \caption{Top: isolated sound board contour (blue) and manually delineated mesh extending over the ribs of the violin (purple). Bottom: zoom on the contour at the level of the ribs.}
    \label{fig:after_isolation}
\end{figure}

\noindent We first pre-process our mesh by manually removing the neck in the MeshLab\footnote{\url{https://www.meshlab.net/}} editing software. Then, when only the body of the instrument remains, we orient it with respect to the principal axes of the frame using Principal Component Analysis (PCA). This aims to align the (approximate) plane of symmetry (more on this in Section \ref{Symmetry plane between the sound board and back}) between the sound board and the back with the $Oxy$ plane (i.e. orthogonal to the $z$ axis) and the left-right plane of symmetry with the $Oxz$ plane (i.e. orthogonal to the $y$ axis). Finally, we manually delimit a mesh that roughly represents the sound board or back. In addition to the sound board/back we want to delineate, this mesh contains a surface that extends over the lateral parts of the violin (which are called the ribs). We then automatically process this overhanging surface to achieve a refined contour isolation. Figure \ref{fig:after_isolation} illustrates the contour resulting from our algorithm (displayed with the Plotly library \cite{plotly}). In purple we can see the manually delineated mesh extending over the ribs and in blue the refined contour that delimits the sound board. As our method is the same for the sound board and the back surfaces, we will use the general `sound board' denomination for both in the following steps.

\begin{figure}[H]
    \centering
    \begin{subfigure}{0.48\textwidth}
    \centering
    \includegraphics[height = 3 cm]{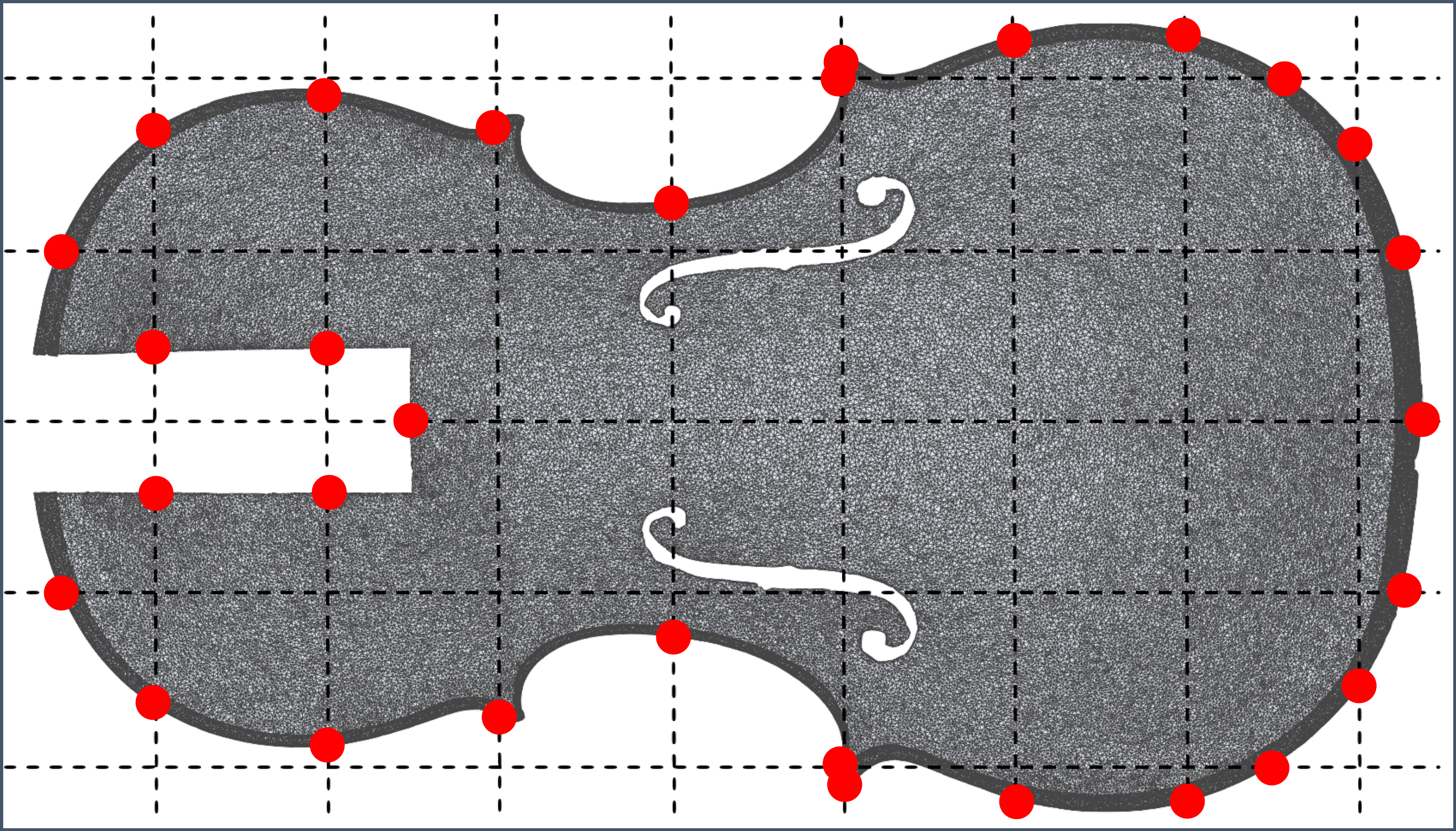}
    \end{subfigure}
    \begin{subfigure}{0.48\textwidth}
    \centering
    \includegraphics[width = 6 cm, height = 3cm]{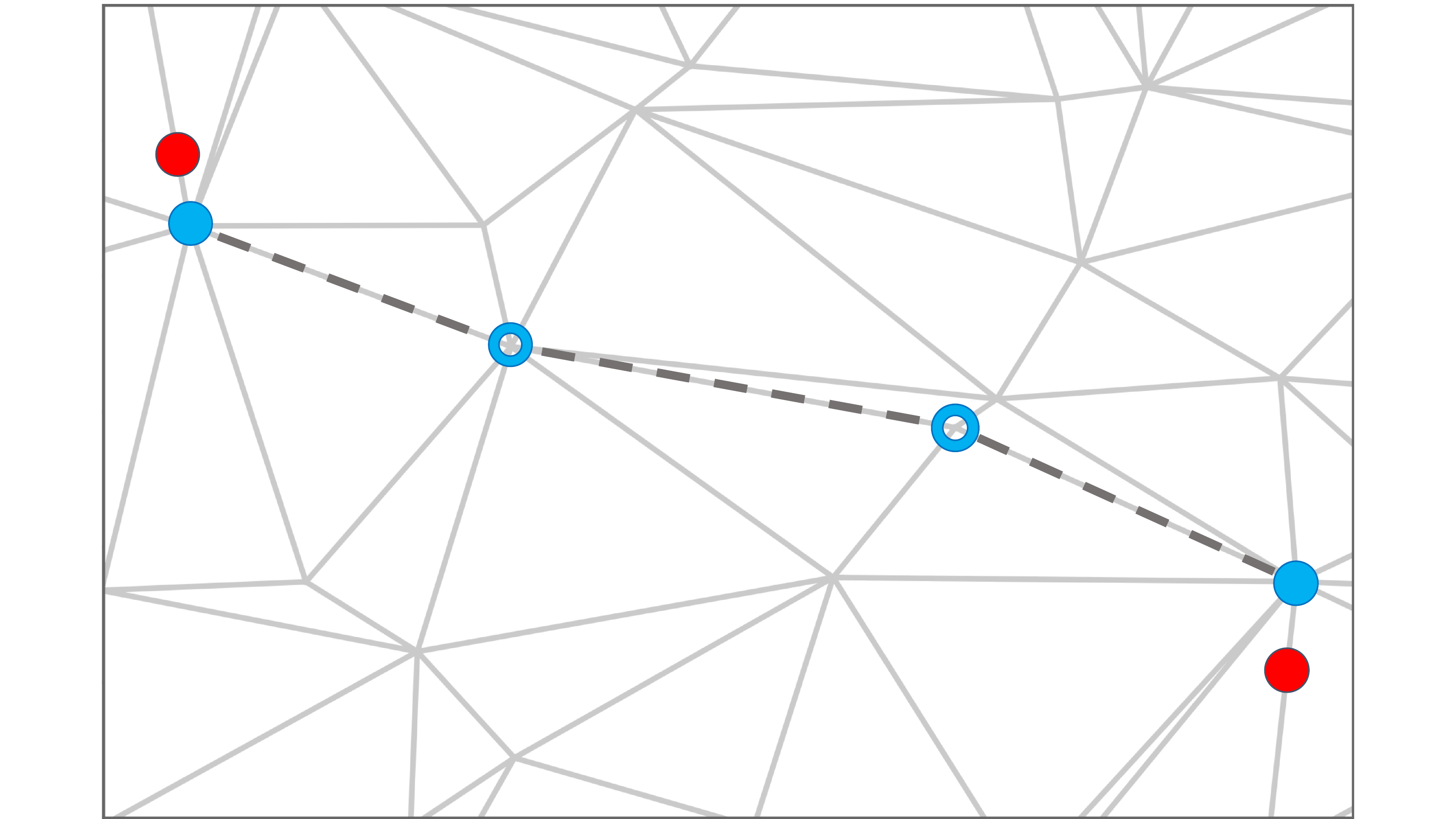}
    \end{subfigure}
    \caption{Contour isolation process. Left: Extreme points computed on a wide grid (Step 1). Right: points and shortest path on the mesh (red: extreme points (Step 1), filled blue: nearest neighbours (Step 2), empty blue: added intermediate vertices (Step 3), dashed gray: shortest paths between two consecutive nearest neighbours (Step 3)).}
    \label{fig:isolation}
\end{figure}

\begin{enumerate}
    \item[] \textbf{Step 1:} We use the orientation of the above-mentioned PCA to compute the `extreme points' that will serve as a starting point for the contour isolation. These extreme points come from vertical cross sections of the sound board every \SI{}{\mm} (orthogonal to either the $x$ or $y$ axis), as we can see in Figure \ref{fig:isolation} (left). We used the Python package \texttt{Meshcut}\footnote{\url{https://github.com/julienr/meshcut}} to compute the planar cuts. These cuts are polylines whose vertices are the intersection between the cutting plane and the edges of the mesh. On each of these cross sections, orthogonal to the horizontal plane, we keep the two most distant points on the cut axis: the extreme points (we make an exception at the level of the manually removed neck to keep more).  \\
    
    \item[] \textbf{Step 2:} The extreme points derived in Step 1 are located on the edges of the original mesh but are not necessarily mesh vertices. Since we want to delineate the sound board by connecting only actual vertices from the initial mesh, we map each extreme point onto their nearest neighbour (NN) on the mesh vertices. Figure \ref{fig:isolation} (right) shows these nearest neighbours. They are computed efficiently with the \textit{Fast Library for Approximate Nearest Neighbours} (FLANN) \cite{muja2009fast}. \\
    
    \item[] \textbf{Step 3:} We then consider the manually delineated mesh as a graph and gather all nearest neighbours into a single closed loop. We first reorder them with a Travelling Salesman Problem (TSP) solver\footnote{\url{https://github.com/dmishin/tsp-solver}} and then link them with a shortest path algorithm\footnote{\url{https://networkx.org/documentation/stable/reference/algorithms/shortest_paths.html}}, using the Euclidean distance as a distance metric between the vertices. If two consecutive nearest neighbours (according to the TSP order) are not adjacent on the graph of the manually delineated mesh, we insert all intermediate vertices between them in the contour in order to obtain a connected loop to isolate the sound board. We can see the added vertices in Figure \ref{fig:isolation} (right). \\
    
    \item[] \textbf{Step 4:} The final step consists in removing the vertices and faces lying `outside' of the closed loop contour determined in Step 3, in order to keep only the `inner mesh’. Once the contour has been calculated, we remove all its vertices and edges from the whole graph, and keep the largest connected component, which corresponds to the `inner mesh' of the sound board.\\
\end{enumerate}

\section{Mesh validation} \label{Validation}

\noindent Here we validate the quality of our photogrammetric meshes by comparing them to the CT scan meshes, and more particularly by comparing their vertices. First, we select a specific metric to register the two point clouds, based on the average distance between corresponding vertices, and show that we obtain a sub-millimetre precision. Then, we compare this metric with other registration techniques present in the literature. Afterwards, we interpret and evaluate the errors that arise from these metrics and finally, we show that we can simplify the mesh to speed up the computation without losing too much accuracy. 

\subsection{Registration between photogrammetric and CT representations}
\label{Registration between photogrammetric and CT representations}
 
\noindent We have now well-defined sound boards and we proceed to quantify the similarity between a photogrammetric and a CT scan mesh of the same instrument. To do so, we compare the corresponding point clouds of the sound boards, whose contours were isolated following the procedure detailed in Section \ref{Contour isolation}. Those representations contain 330k to 480k vertices. We have considered that the CT mesh is a priori the most accurate, and that it will therefore be used as the reference mesh. The classical Hausdorff distance between two sets does not fit our purpose, as it does not quantify the overall similarity, but only focuses on the worst-case distance between corresponding vertices. Instead, following \cite{5687883}, we introduce a specific distance metric between two meshes based on their vertices:
\begin{equation}
    D(s,p) = \frac{1}{N_s}\sum_{i=1}^{N_s}\left\| \bm{s_i} - \bm{p_{nn(i)}}\right\|
    \tag{D}
\label{eq:D}
\end{equation}
where 
\begin{equation*}
    \bm{p_{nn(i)}} = \arg \min_{p_j \ \in \ p} \left\| \bm{s_i} - \bm{p_j} \right\|
    \tag{NN}
\label{eq:NN}
\end{equation*}

\begin{figure}[H]
    \centering
    \begin{subfigure}{0.48\textwidth}
    \centering
    \includegraphics[width = 6 cm, valign = t]{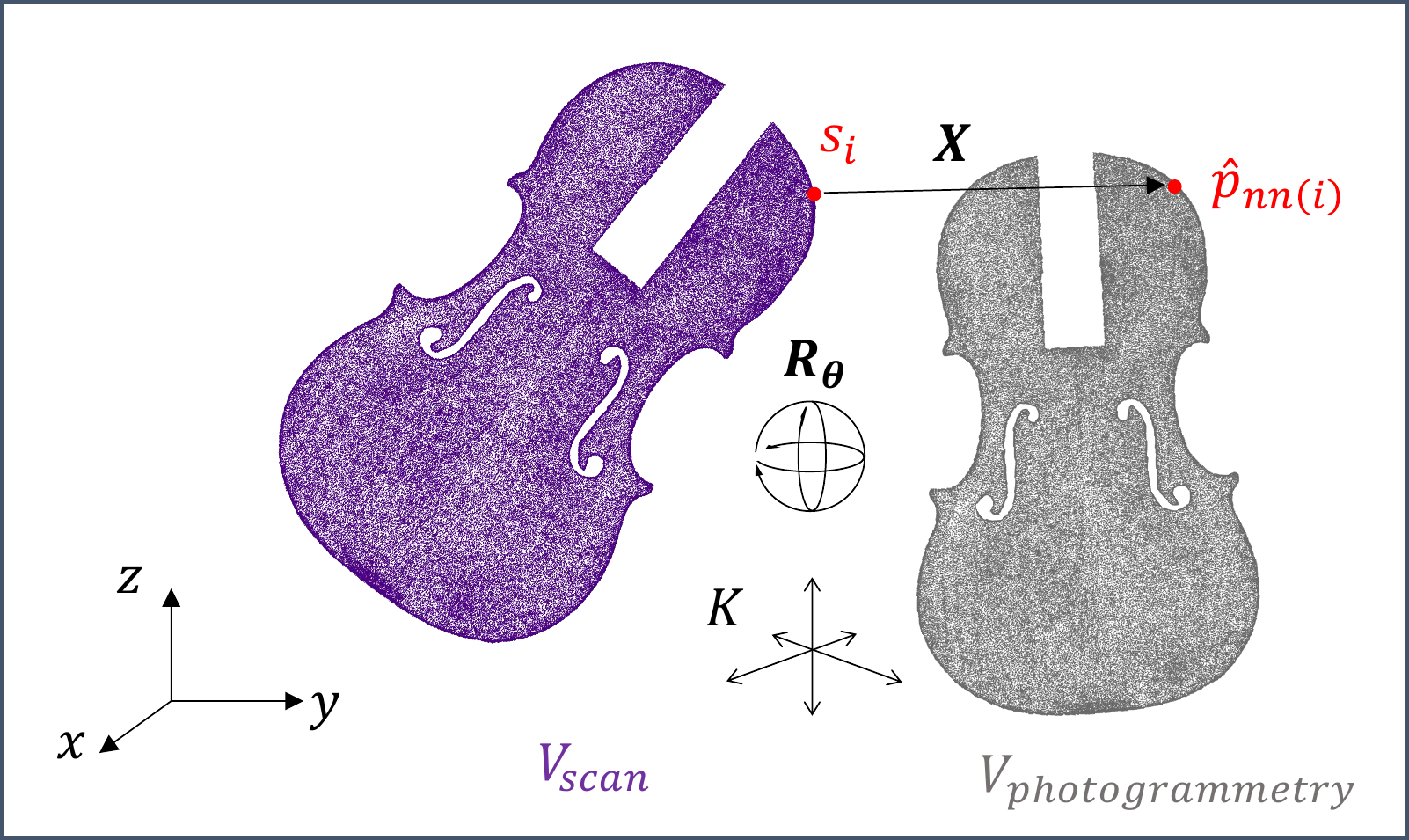}
    \end{subfigure}
    \begin{subfigure}{0.48\textwidth}
    \centering
    \includegraphics[width = 6.25 cm, valign = t]{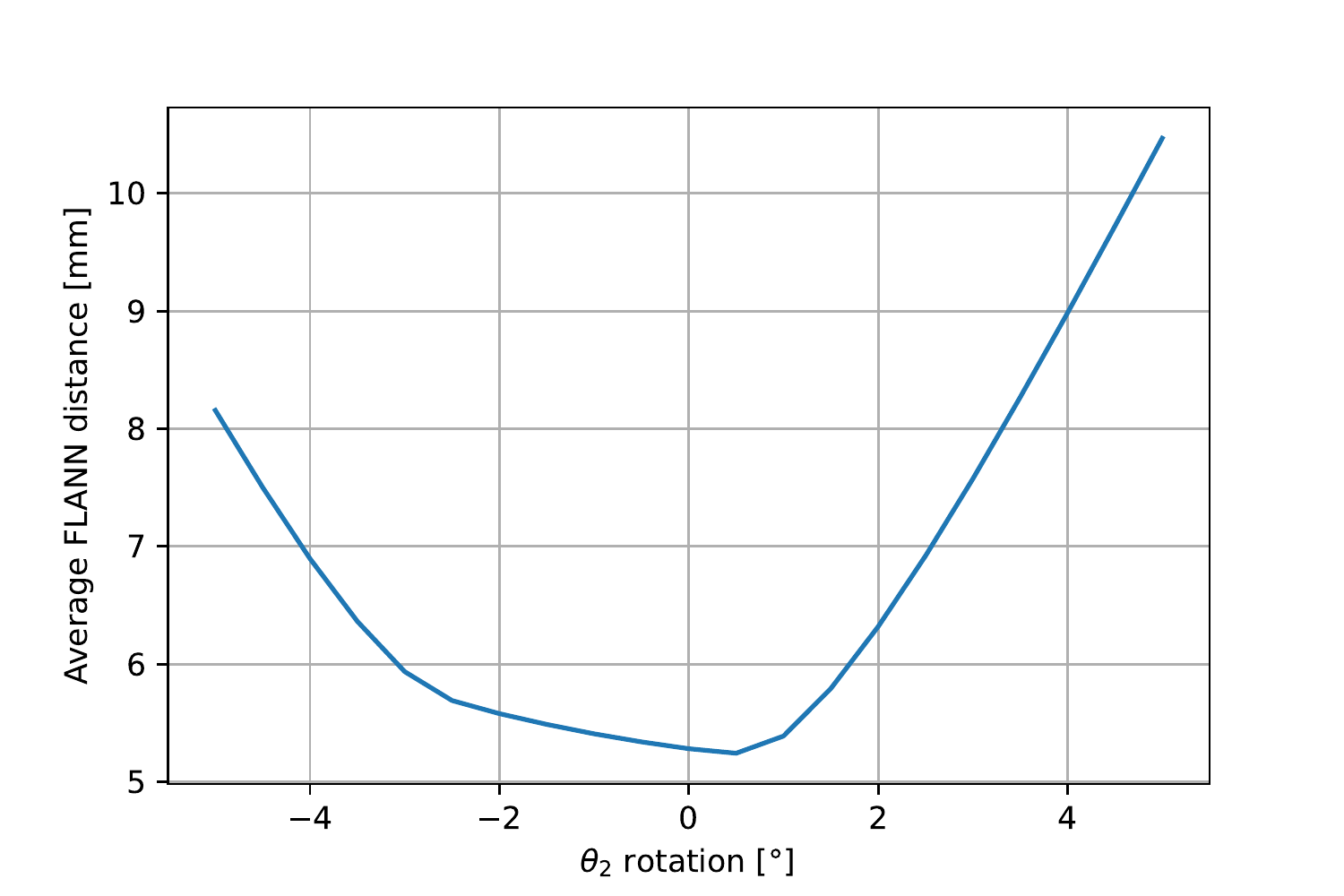}
    \end{subfigure}
    \caption{Matching problem (left) and average distance $D$ when varying a single angle $\theta_2$ before optimisation (right).}
    \label{fig:opti}
\end{figure}

\noindent This metric \ref{eq:D} is the average Euclidean distance between each vertex $\{ \bm{s_{i}} \}_{i=1,\ldots, N_s}\in \mathbb{R}^3$ of the CT cloud $\bm{s}$ and its nearest neighbour $\{ \bm{p_{nn(i)}} \}_{i=1,\ldots, N_s}\in \mathbb{R}^3$ from the photogrammetric cloud $\bm{p}$. However, as the two point clouds are not aligned, we need first to identify the optimal translation, rotation and scaling factor that produce the minimum average distance as defined in \ref{eq:D}. We therefore optimise the seven parameters $\bm{X}\in\mathbb{R}^3$, $\bm{\theta}\in \mathbb{R}^3$ and $K \in \mathbb{R}$ describing the translation, rotation and scaling\footnote{As the scaling has been previously performed manually (see Section \ref{Photogrammetric mesh}), we insert this factor $K$ to correct the potential error induced by this operation.} that the photogrammetric point cloud has to undergo in order to best match the CT point cloud, as in Figure \ref{fig:opti} (left), and we solve:
\begin{equation}
    \min_{(\bm{X},\bm{R_{\theta}},K)} D\left(s,\hat{p}\left(\bm{X},\bm{R_{\theta}},K\right)\right)
    \tag{MinD}
\label{eq:Min_D}
\end{equation}
where, for each vertex $\{ \bm{p_{j}} \}_{j=1,\ldots, N_p} \in \mathbb{R}^3$ of the photogrammetric cloud $\bm{p}$, a rigid body transformation \ref{eq:RBT} is applied:
\begin{equation*}
    \bm{\hat{p}_j} = K \left( \bm{R_\theta}\bm{p_j} + \bm{X} \right)
    \tag{RBT}
\label{eq:RBT}
\end{equation*}
with $\bm{R_\theta}$ the rotation operator for a rotation sequence $\theta_1 \rightarrow \theta_2 \rightarrow \theta_3$
\[
\small
\hspace*{-.15cm}
\bm{R_\theta} = \left(\begin{array}{ccc}
\cos \theta_3 & \sin \theta_3 & 0 \\
-\sin \theta_3 & \cos \theta_3 & 0 \\
0 & 0 & 1
\end{array}\right)
\left(\begin{array}{ccc}
\cos \theta_2 & 0 & -\sin \theta_2 \\
0 & 1 & 0 \\
\sin \theta_2 & 0 & \cos \theta_2
\end{array}\right)
\left(\begin{array}{ccc}
1 & 0 & 0 \\
0 & \cos \theta_1 & \sin \theta_1 \\
0 & -\sin \theta_1 & \cos \theta_1
\end{array}\right).
\]

\noindent In simpler terms, \ref{eq:Min_D} is the minimum distance between the CT scan mesh and the photogrammetric mesh, which has undergone a rigid body transformation to best match the position, orientation and size of the CT scan mesh. In Equation \ref{eq:NN}, we still compute efficiently the nearest neighbour of each of the $N_s$ vertices $ \bm{s_{i}}$ among the $N_p$ (later transformed) vertices $\bm{p_{j}}$ with the FLANN. The minimisation \ref{eq:Min_D} is performed with the Powell method \cite{powell1964efficient} and its Python implementation \texttt{scipy.optimize.fmin\textunderscore powell}\footnote{We set an objective function tolerance of $10^{-5}$ as a convergence criterion.}. The total computation time is about one hour on a standard laptop. Both clouds were first oriented using the principal axes from Principal Component Analysis (PCA), while their relative positions and scaling were adjusted manually (see Sections \ref{Photogrammetric mesh} et \ref{CT scan mesh} for the scaling). Results of the matching problem are presented in Table \ref{tab:MinProb}. 
\begin{table}[H]
    \centering
    \begin{tabular}{|c|c|c c c c|c|}
    \hline
         Violins & Average distance $D$ & $\phantom{-}\theta_1$ & $\phantom{-}\theta_2$ & $\phantom{-}\theta_3$ & K  \\
         \hline
         \textbf{Hofmans} & 0.301 & $-0.261$ & $-0.794$ & $-0.486$ & $1.024$ \\
         \textbf{Cuypers} & 0.215 & $-0.050$ & $\phantom{-}0.085$ & $\phantom{-}0.024$ & 1.029 \\
         \hline
    \end{tabular}
    \caption{Average distance [\SI{}{mm}] between the CT and photogrammetric sound board clouds, optimal angles $[\degree]$ and scaling factor $K$ $[/]$}
    \label{tab:MinProb}
\end{table}

\noindent We achieve an average sub-millimetre accuracy between both representations. Angles in the optimal alignment range from $\ang{0.024}$ to $\ang{0.794}$, indicating that the initial orientation obtained from PCA was relatively accurate, especially since the distance $D$ is very sensitive to the orientation of the angles. We see in Figure \ref{fig:opti} (right) that when we vary the value of a single angle (here $\theta_2$, before optimisation), this distance increases.

\subsection{Alternative registrations}
\label{Alternative registrations}

\begin{figure}[H]
    \centering
    \includegraphics[scale=0.42]{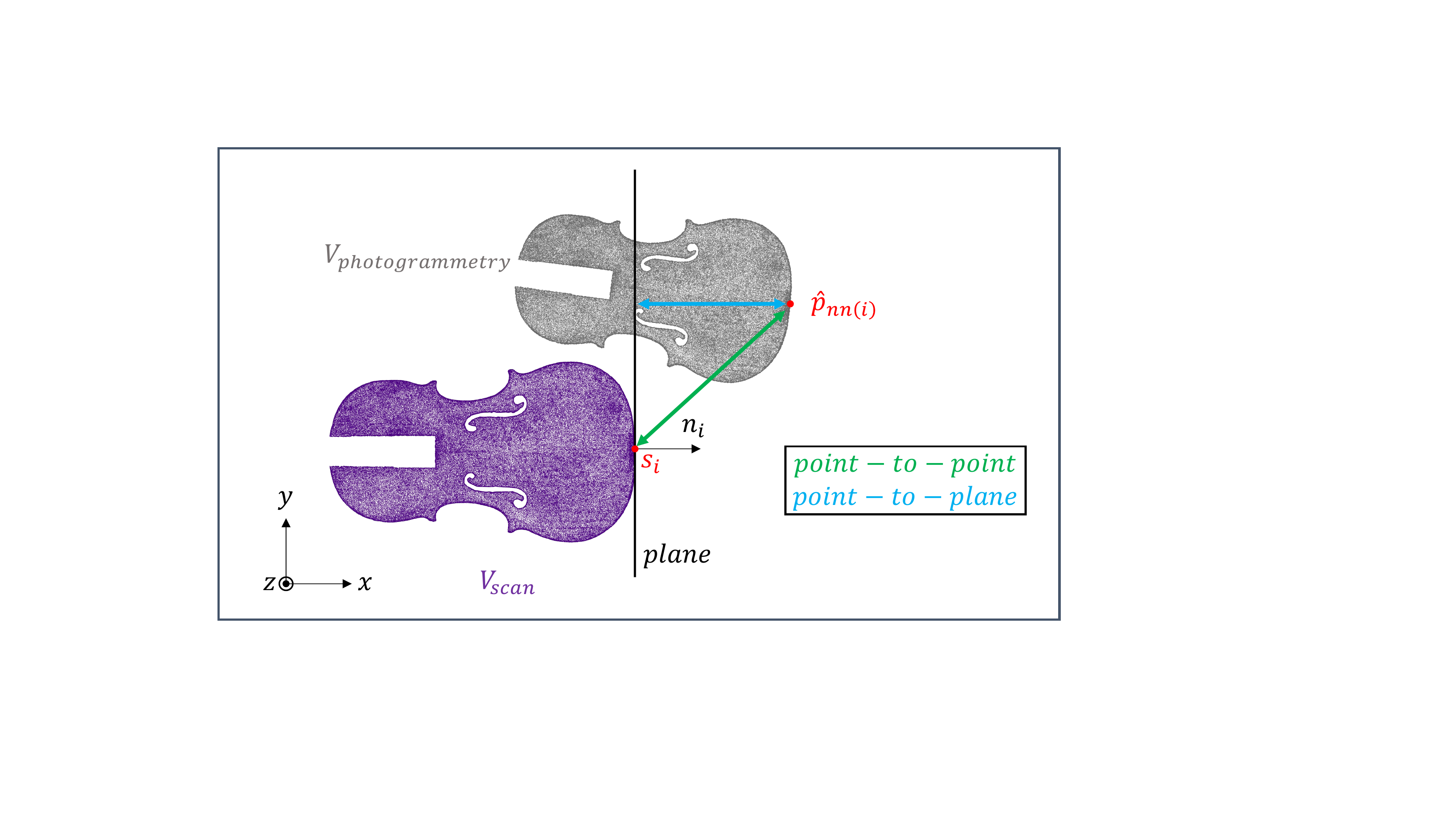}
    \caption{Comparison between point-to-point and point-to-plane approaches.}
    \label{fig:point_to_point/plane}
\end{figure}

\noindent An alternative to our approach is to directly use a point cloud registration algorithm, such as the Iterative Closest Point (ICP) \cite{chen1992object,besl1992method}. This algorithm has the advantage of being faster, but as it only optimises the position of a subsample of the points on the whole cloud, it could be less accurate than our method described in Section \ref{Registration between photogrammetric and CT representations}. We compared the accuracy obtained with our method and the \texttt{SimpleICP} implementation\footnote{\url{https://github.com/pglira/simpleICP}} proposed in \cite{glira2015a}. Rather than using a classical point-to-point distance \cite{besl1992method} between corresponding vertices, the ICP algorithm uses a point-to-plane distance \cite{chen1992object}, whose convergence has proven to be faster \cite{924423}. The squared point-to-plane distance between two meshes is:

\begin{equation}
    D^2_{plane}(s,p) = \frac{1}{N_s}\sum_{i=1}^{N_s} \left| (\bm{s_i}-\bm{p_{nn(i)}})^T \cdot \bm{n_i} \right| ^2
    \tag{$D^2_\mathrm{plane}$} 
     \label{eq:D_plane}
\end{equation}

with $\bm{p_{nn(i)}}$ the nearest neighbour of each vertex $\bm{s_i}$ as defined in \ref{eq:NN} and $\bm{n_i}$ the normal vector of each vertex $\bm{s_i}$ of the CT scan mesh. The normal vector of each vertex can be estimated using a principal component analysis of the covariance matrix of the coordinates of neighbouring points \cite{glira2015a, shakarji1998least}. Figure \ref{fig:point_to_point/plane} illustrates the difference between the two error metrics. Incidentally, it shows that point-to-plane is always smaller than point-to-point.

\noindent Once again, we want to minimise this distance and therefore optimise the seven parameters of the rigid body transformation that the photogrammetric point cloud has to undergo to best match the CT point cloud. We then solve:

\begin{equation}
    \min_{(\bm{X},\bm{R_{\theta}},K)} D^2_{plane}\left(s,\hat{p}\left(\bm{X},\bm{R_{\theta}},K\right)\right)
    \tag{$MinD^2_\mathrm{plane}$}
     \label{eq:Min_D_plane}
\end{equation}
with $\hat{p}(\bm{X}, \bm{R_{\theta}}, K)$ the photogrammetric mesh after a rigid body transformation \ref{eq:RBT}. \\

\noindent We notice that there is no scaling factor $K$ in the \texttt{SimpleICP} implementation, and thus tried a second ICP run after applying a fixed external scaling factor (obtained with our point-to-plane implementation). In addition, as the point-to-plane approach minimises an average of square distances, we also tried to minimise the point-to-point approach adding a square term in Equation \ref{eq:D}, which is nothing more than minimising the Mean Square Error (MSE) of the errors distribution: 

\begin{equation}
    D^2(s,p) = \frac{1}{N_s}\sum_{i=1}^{N_s}\left| \bm{s_i} - \bm{p_{nn(i)}}\right|^2
    \tag{$D^2$}
\label{eq:D^2}
\end{equation}
and 
\begin{equation}
    \min_{(\bm{X},\bm{R_{\theta}},K)} D^2\left(s,\hat{p}\left(\bm{X},\bm{R_{\theta}},K\right)\right).
    \tag{$MinD^2$}
\label{eq:Min_D2}
\end{equation}
with $\hat{p}(\bm{X}, \bm{R_{\theta}}, K)$ the photogrammetric mesh after a rigid body transformation \ref{eq:RBT}. \\

\noindent The results of optimising all the objective functions described above are shown for both violins in Tables \ref{tab:MinProb_all_Hofmans} and \ref{tab:MinProb_all_Cuypers}. The left-hand column shows which metric was used to calculate the optimal parameters and each row displays the value of the three considered metrics $\left( D, D^2 \text{ and } D^2_{plane} \right)$ for each set of optimised parameters. Note that a square root is applied to the $D^2$ and $D^2_{plane}$ metrics to obtain a distance in \SI{}{\mm}, and to be more comparable to the $D$ metric.

\begin{table}[H]
    \centering
    \begin{tabular}{|l|c|c|c|}
    \hline
          Optimised metrics \hspace{0.2 cm} \textbackslash \textbackslash \textbackslash \hspace{0.2 cm} Displayed metrics & $D$ & $\sqrt{D^2}$ & $\sqrt{D^2_{plane}}$ \\
         \hline
         Point-to-point $D$ & \textbf{0.301} & 0.363 &  0.262 \\
         Point-to-point square $D^2$ & 0.302 & \textbf{0.359} & 0.261 \\
         Point-to-plane $D^2_{plane}$ (ours, with scaling) & 0.304 & 0.362 & \textbf{0.258} \\
         Point-to-plane $D^2_{plane}$ (ICP, external scaling) & 0.302 & 0.363 & 0.264 \\
         Point-to-plane $D^2_{plane}$ (ICP, no scaling) & 0.474 & 0.796 & 0.650 \\
         \hline
    \end{tabular}
    \caption{Optimal distances [\SI{}{mm}] for Hofmans' violin. Left column: metric with which the parameters were computed.\\ Top row: metric for which we computed distances with the optimised parameters of the left column\\ (point-to-point, point-to-point square and point-to-plane respectively).}
    \label{tab:MinProb_all_Hofmans}
\end{table}

\begin{table}[H]
    \centering
    \begin{tabular}{|l|c|c|c|}
    \hline
          Optimised metrics \hspace{0.2 cm} \textbackslash \textbackslash \textbackslash \hspace{0.2 cm} Displayed metrics & $D$ & $\sqrt{D^2}$ & $\sqrt{D^2_{plane}}$ \\
         \hline
         Point-to-point $D$ &\textbf{ 0.215} & 0.265 &  0.125 \\
         Point-to-point square $D^2$ & 0.217 & \textbf{0.263} & 0.130 \\
         Point-to-plane $D^2_{plane}$ (ours, with scaling) & 0.216 & 0.270 & \textbf{0.122} \\
         Point-to-plane $D^2_{plane}$ (ICP, external scaling) & 0.215 & 0.267 & 0.123 \\
         Point-to-plane $D^2_{plane}$ (ICP, no scaling) & 0.633
 & 1.164 & 0.912 \\
         \hline
    \end{tabular}
    \caption{Optimal distances [\SI{}{mm}] for Cuypers' violin. Left column: metric with which the parameters were computed.\\ Top row: metric for which we computed distances with the optimised parameters of the left column\\ (point-to-point, point-to-point square and point-to-plane respectively).}
    \label{tab:MinProb_all_Cuypers}
\end{table}

\noindent First, we observe that all our errors are sub-millimetre, which again confirms that our photogrammetric approach is accurate. Furthermore, we see that the point-to-point distance $D$ (arithmetic mean of the errors) is better than the squared point-to-point distance metric $\sqrt{D^2}$ (quadratic mean of the errors, also called RMSE). This also makes sense because optimising the (R)MSE is more sensitive to outliers than optimising the arithmetic mean of the errors, and we know from the Arithmetic Mean (AM)--Quadratic Mean (QM) Inequality that $AM \leq QM$. \\ 

\noindent Then, we focus on the point-to-plane distance $\sqrt{D^2_{plane}}$ (expressed in \SI{}{\mm}). We first see that, as expected, it is smaller than the point-to-point square distance $\sqrt{D^2}$, also in \SI{}{mm} (RMSE). We compared our approach, which allows scaling, to \texttt{SimpleICP}, which does not, and immediately see that this $K$ factor greatly improves the results. Nevertheless, when we provide an external scaling factor to \texttt{SimpleICP}, we see that the two results become quite comparable. \texttt{SimpleICP} has the advantage of being faster than our method because it is based on a sample of points and not the whole cloud. On the other hand, it does not allow scaling, which was necessary in our case. Moreover, this algorithm is based on `artificial' normals $\bm{n_i}$, an information initially absent from our mesh. In any case, we have performed an exhaustive validation. \\

\noindent Are the above metrics the best for interpreting the results? We have used two types of approaches: point-to-point which matches a point with a point of the other cloud, and point-to-plane which matches a point with an `infinite' plane of the other cloud. A proposal for a third metric would be to compute the point-to-face distance that matches a point with a face of the other mesh, namely a `finite' planar section (i.e. a polygon). It is easy to see that this metric will always lie between the other two, and would probably provide a more intuitive definition of the distance between two meshes. Moreover, it would include the use of faces, which are explicit elements of the mesh but are not used in the point-to-point metric. However, projecting a large number of points onto a corresponding face appears to be too computationally expensive for this intent. \\

\noindent Each metric in Tables \ref{tab:MinProb_all_Hofmans} and \ref{tab:MinProb_all_Cuypers} appears to be the smallest for its optimisation criterion, as expected. Interestingly, the four methods considered with scaling provide almost identical results, e.g. optimising the point-to-plane distance almost provides the optimal result for the point-to-point distance, and vice versa. The optimised angles and scaling (not shown here) vary very little from one metric to another. As the metrics provide similar results in the final point cloud registration, we chose to continue working with the point-to-point distance $D$ that was introduced first. 

\subsection{Error assessment and validation}
\label{Error assessment and validation}
\begin{figure}[H]
\centering
\begin{subfigure}[b]{0.48\textwidth}
\includegraphics[height = 4.6 cm]{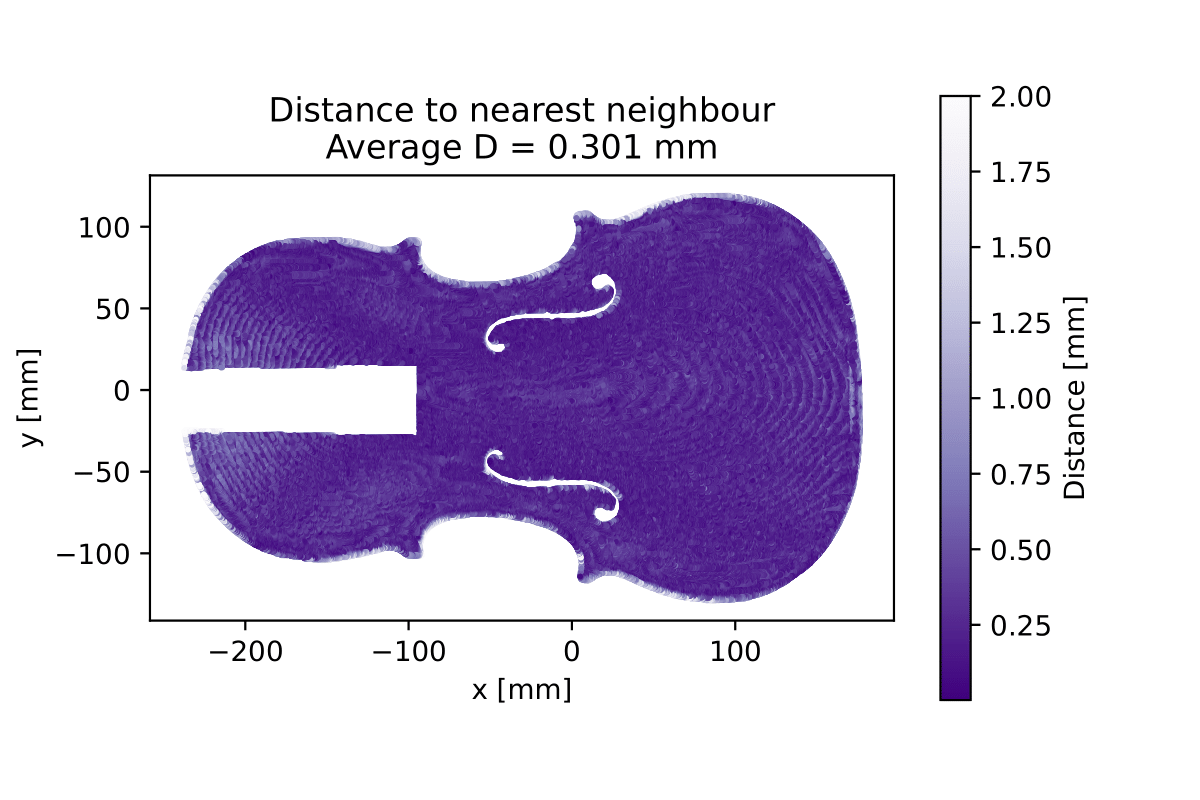}
\vspace{0.01\baselineskip}
\includegraphics[height = 3.9 cm]{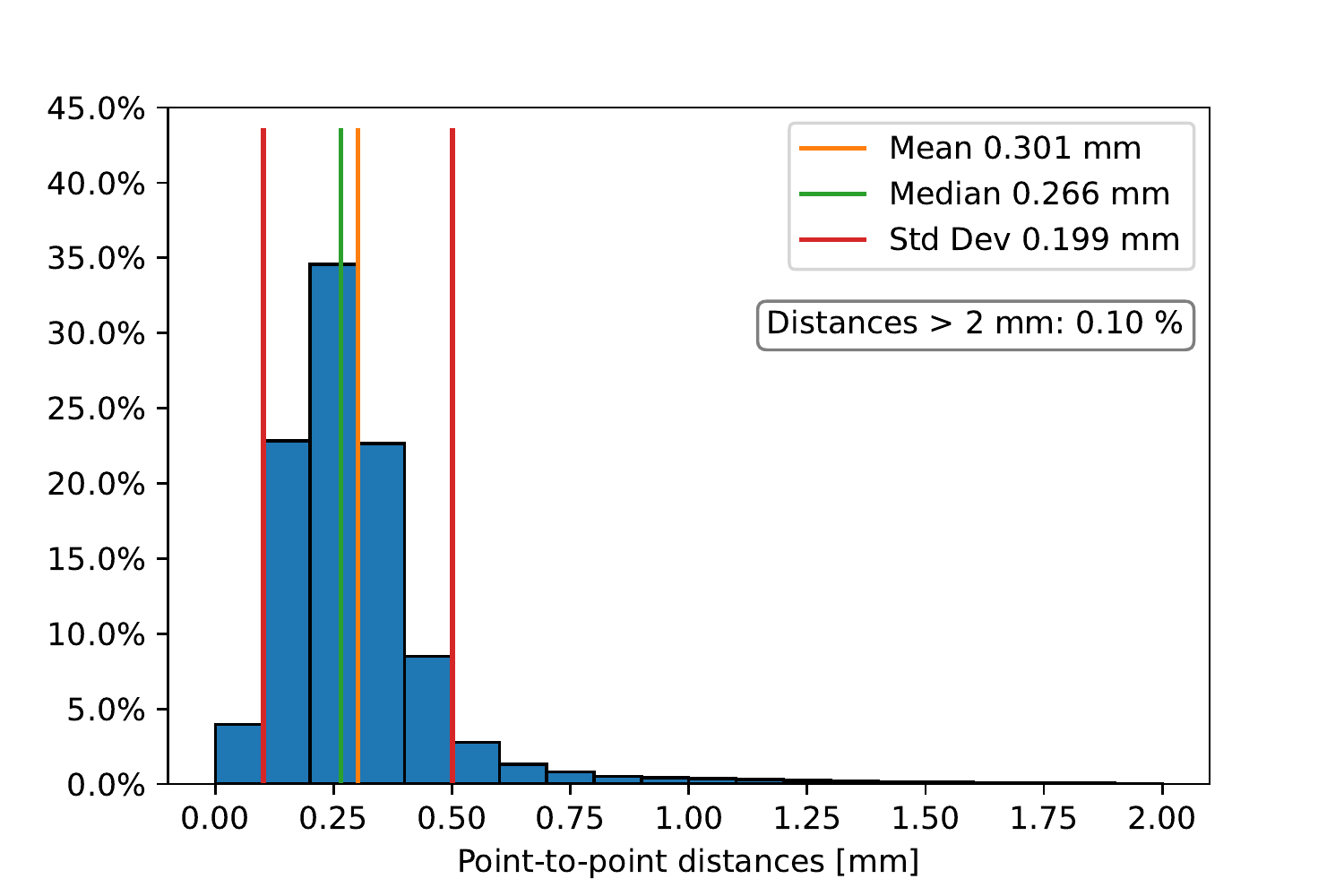}
\end{subfigure}%
\begin{subfigure}[b]{0.48\textwidth}
\includegraphics[height = 4.6 cm]{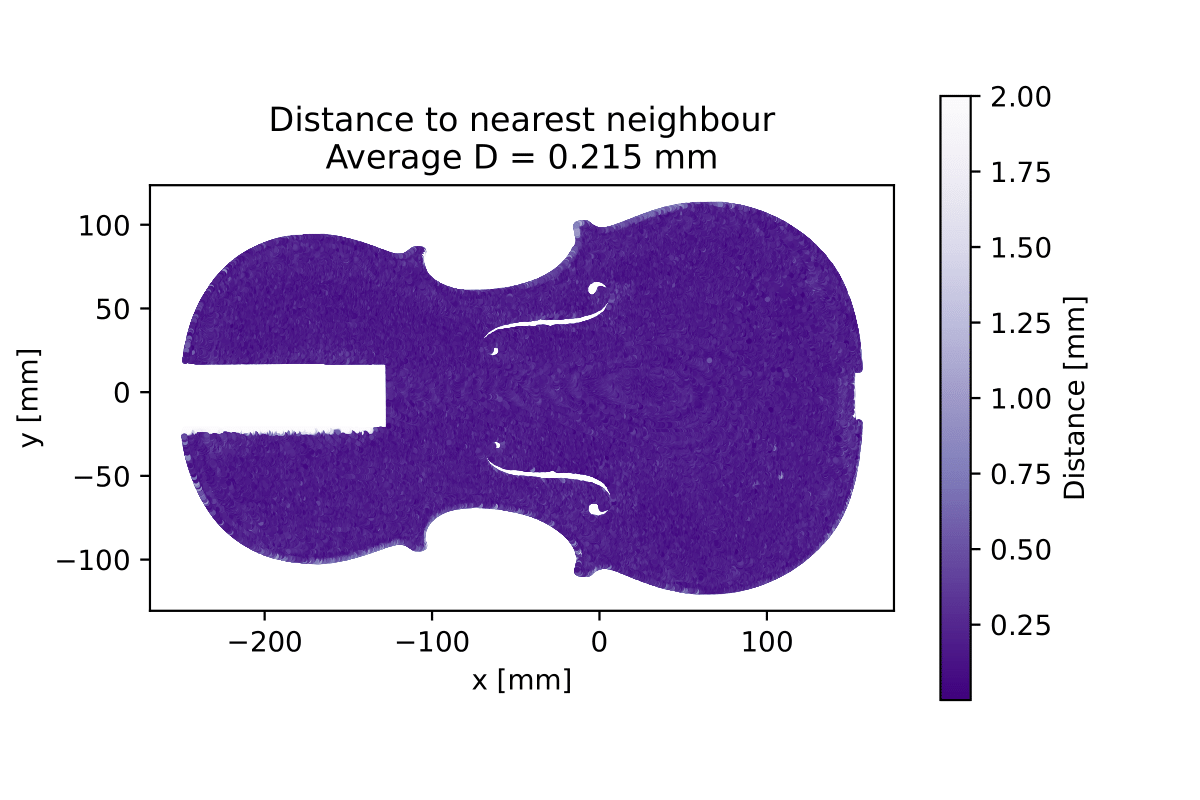}
\vspace{0.01\baselineskip}
\includegraphics[height = 3.9 cm]{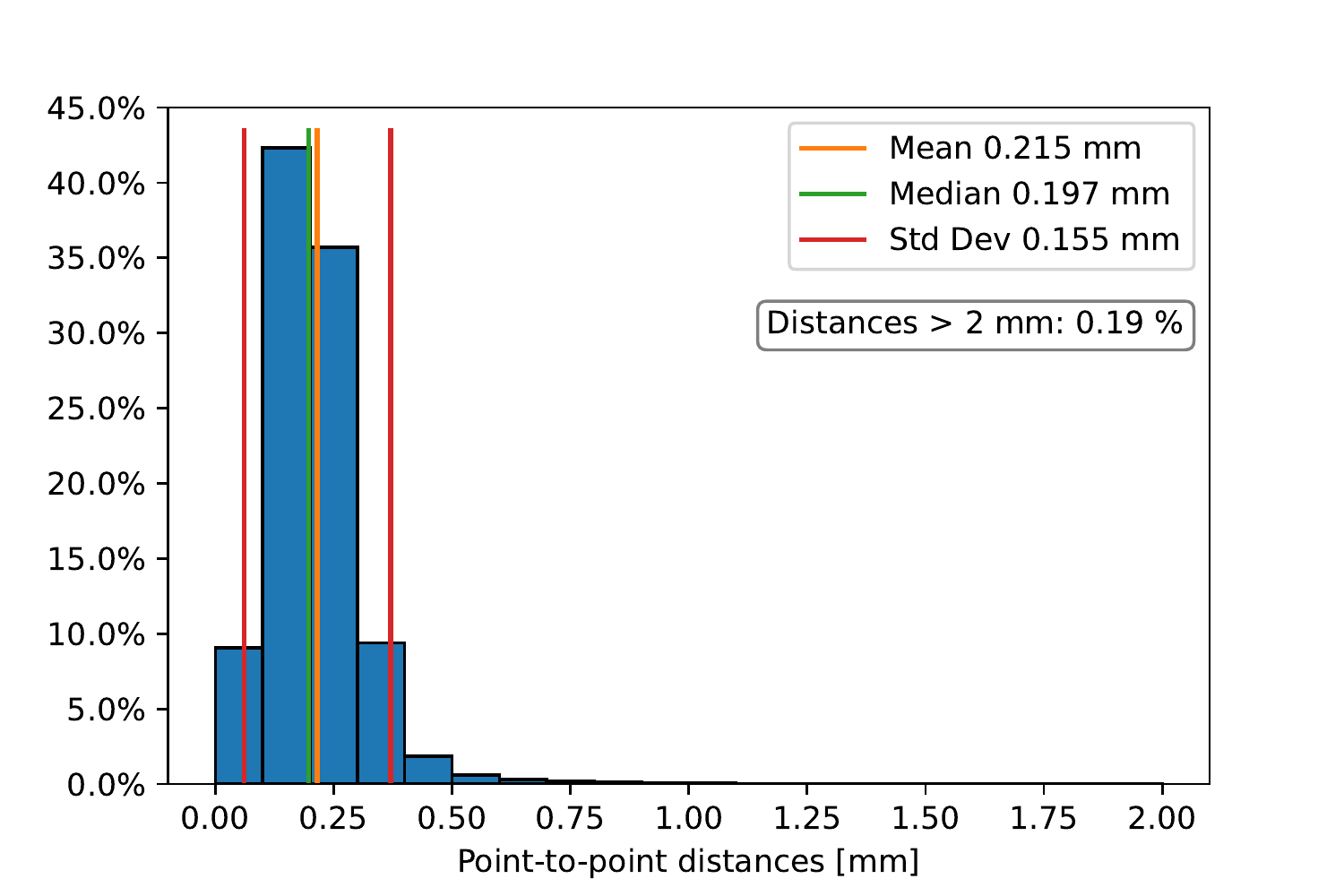}
\end{subfigure}%
\caption{Distribution of point-to-point distances [\SI{}{mm}] from CT point cloud to the nearest neighbour in photogrammetric cloud (left: Hofmans, right: Cuypers).}
\label{fig:Data}
\end{figure}

\noindent The average error between the CT and photogrammetric point clouds lies in the sub-millimetre range for both instruments, which is rather small. The distribution of the point-to-point distances between vertices of the CT mesh to the nearest photogrammetric vertices can be observed using heat maps and histograms in Figure \ref{fig:Data} (left: Hofmans, right: Cuypers), showing very good agreement throughout the sound boards, and very few distances larger than \SI{2}{\mm} (respectively $0.10 \%$ and $0.19 \%$ for the Hofmans and Cuypers instruments). We conclude from the small average errors in Table \ref{tab:MinProb} and from our histograms and heat maps in Figure \ref{fig:Data} that our photogrammetric approach with respect to medical scans is validated. More comparisons to strengthen our validation can be found in \ref{Appendix}.

\subsection{Simplification}
\label{Simplification}

\noindent The meshes validated in Sections \ref{Registration between photogrammetric and CT representations}, \ref{Alternative registrations} and \ref{Error assessment and validation} are very dense and therefore will slow down all calculations we perform on them. Before turning to the geometric analysis of the instruments in Section \ref{Geometric analysis}, we study a simplification procedure that would offer a trade-off between computational speed and accuracy. \\

\noindent MeshLab includes a simplification process based on the Quadric Edge Collapse Decimation algorithm \cite{garland1997surface}. Very generally, the algorithm iteratively calculates the contraction of vertex pairs which cause the least possible error (with a quadric error metric), contracts the minimum cost pair and repeats. The procedure ends when a prespecified number of faces is reached. Concretely, for two given vertices $v_1$ and $v_2$ on the same edge to be contracted, the new vertex $\bar{v}$ lies somewhere on that edge connecting $v_1$ and $v_2$. Thus, a simplified mesh no longer shares exactly the same vertices as the original mesh. \\

\begin{figure}[H]
    \centering
    \begin{subfigure}{0.48\textwidth}
    \centering
    \includegraphics[height = 4.7 cm]{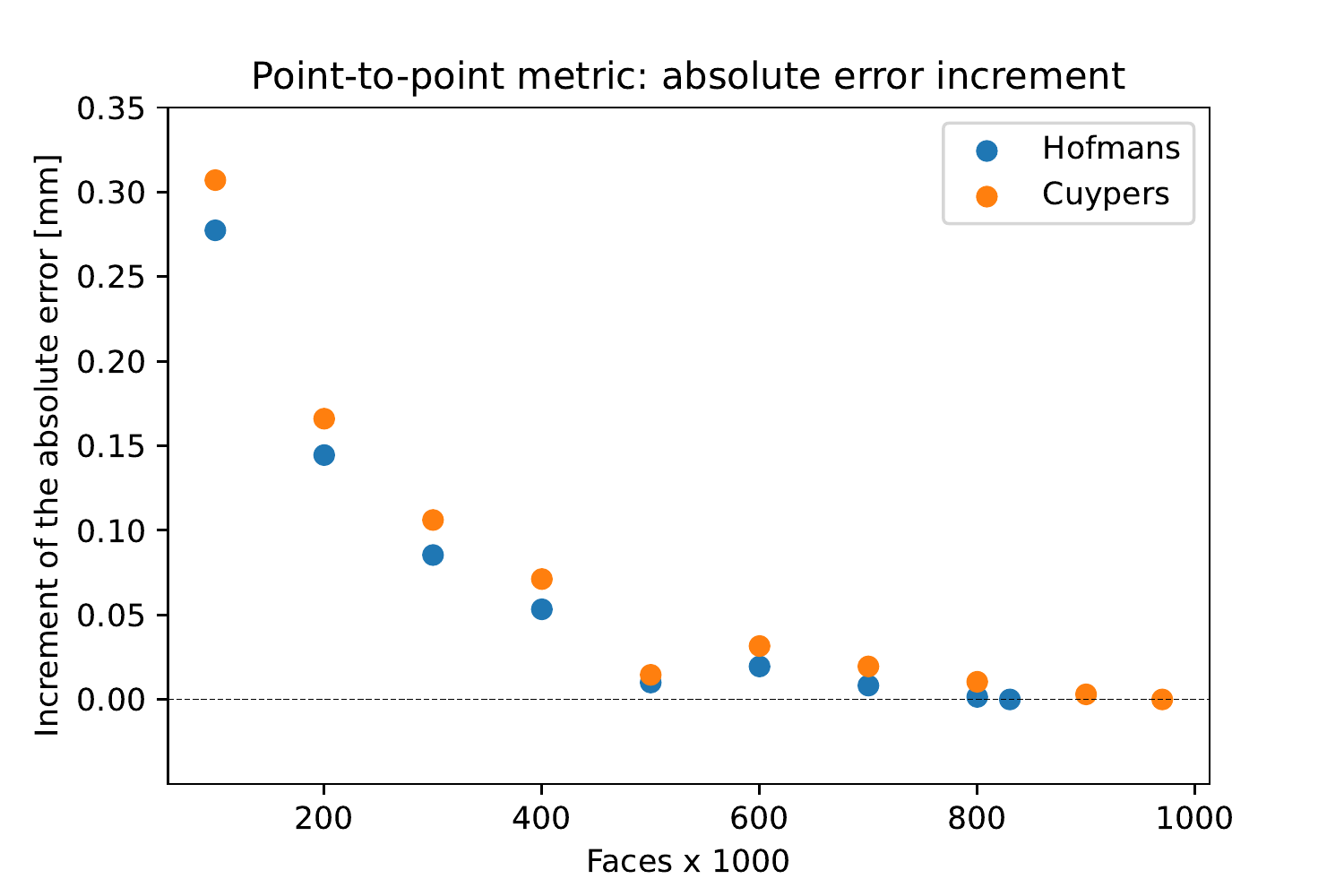}
    \end{subfigure}
    \hfill
    \begin{subfigure}{0.48\textwidth}
    \centering
    \includegraphics[height = 4.7 cm]{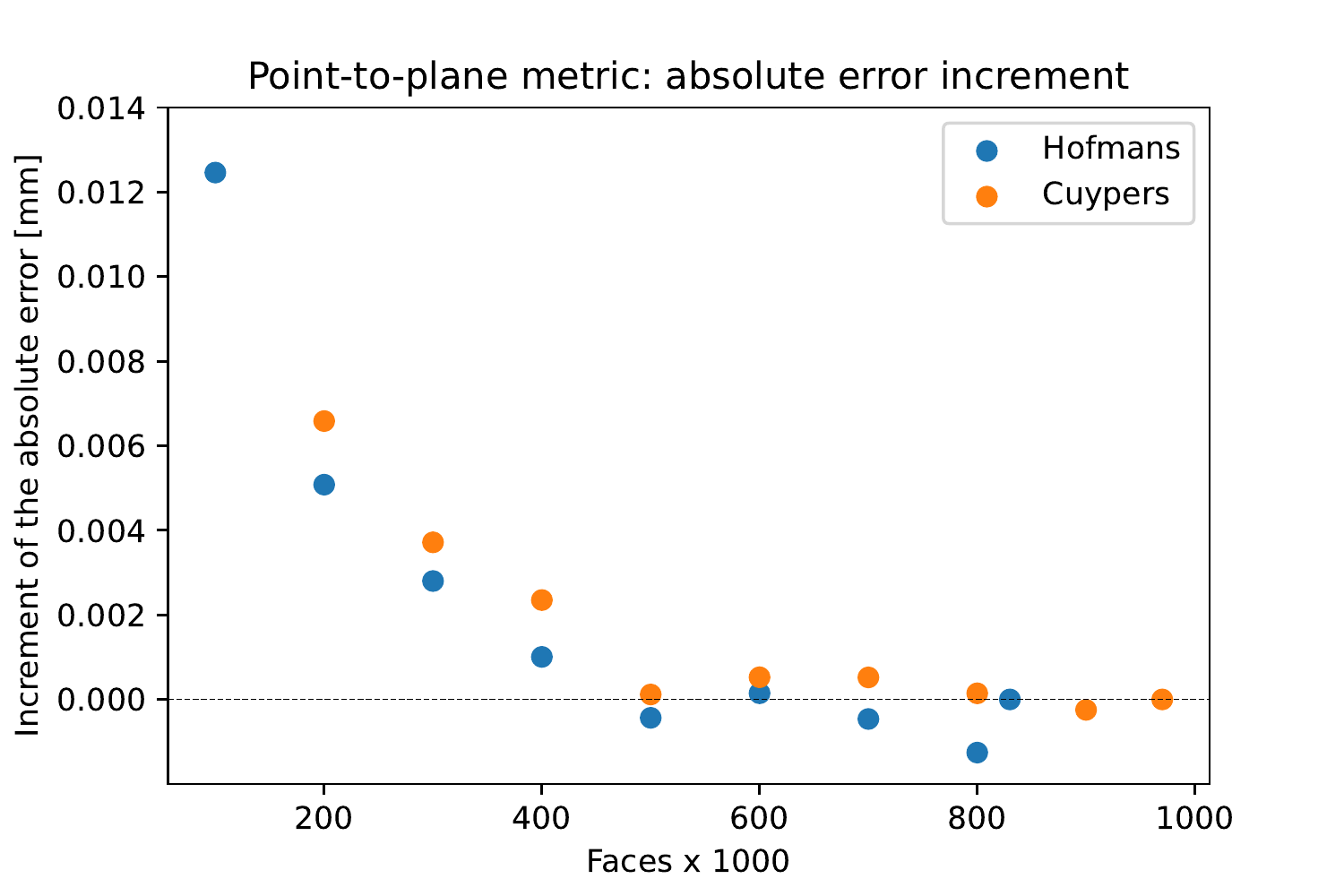}
    \end{subfigure}
    \caption{Increment of the absolute error with respect to the original sound board meshes for point-to-point (left) and point-to-plane metrics (right).}
    \label{fig:absolute error}
\end{figure}

\noindent Figure \ref{fig:absolute error} shows the increment in absolute error for the point-to-point (left) and point-to-plane (right) metrics when comparing the CT scan mesh and several versions of the simplified photogrammetric meshes (in terms of number of faces). The error between the CT mesh and the original full photogrammetric mesh is considered to be the reference, i.e. corresponds to an increment of \SI{0}{mm} (also displayed as horizontal dashed line).

\noindent Noticeably, the relationship between the error increment and the number of faces is not linear. It seems that a simplification to 500k faces (about 250k vertices) offers a good trade-off between accuracy and computational effort. We note that these point-to-point and point-to-plane error metrics (see Sections \ref{Registration between photogrammetric and CT representations} and \ref{Alternative registrations}) are extremely sensitive to the exact position of the vertices, which is repeatedly modified by the Quadric Edge Collapse Decimation algorithm. We therefore suggest to use another metric to assess the impact of the simplification process. \\

\noindent We propose to linearly interpolate the meshes on a regular horizontal grid (in the sense of the PCA plane of the CT violin), with nodes equally spaced every $\SI{1}{mm} \times \SI{1}{mm}$. Thus, for each node on the grid, we draw a vertical normal, identify the face through which this normal intersects the mesh and calculate its $z-$coordinate based on the 3-dimensional plane equation of the intersected face. We compared the vertical differences between the Cuypers CT scan sound board and the original photogrammetric mesh of the sound board (830k faces, 420k vertices) then its simplifications of 250k and 50k vertices respectively. We used the parameters of the rigid body transformation found in Section \ref{Registration between photogrammetric and CT representations} (see Table \ref{tab:MinProb}) to make the three photogrammetric sound boards match the reference CT sound board before calculating the vertical differences, which are summarised in Table \ref{tab:Vertical_dist_CT}. As we see almost no difference between the CT scan and the photogrammetric simplifications, we decided to compare these simplifications to each other, using a better accuracy (see Table \ref{tab:Vertical_dist_photogrammetry}). We see that in contrast to the point-to-point criterion, the simplification leads to a very small vertical error. This made us reconsider the observable kink in Figure \ref{fig:absolute error}, at the level of the 500k faces. It may be due to the enlargement of the mesh faces (triangles) that occurs when simplifying the photogrammetric meshes, rather than actually indicating a poor quality mesh. Considering this new metric, which is certainly more relevant to study the impact of the simplification, we decided to select photogrammetric meshes of 100k vertices to analyse the geometry of instruments in Section \ref{Geometric analysis}.

\begin{table}[H]
    \centering
    \begin{tabular}{|l|c|c|c|}
    \hline
          Cuypers sound board & 420k vertices & 250k vertices & 50k vertices \\
         \hline
         Maximum distance & 1.94 & 1.94 & 1.93 \\
         Mean & 0.091 & 0.091 & 0.091 \\
         Median & 0.079 & 0.079 & 0.079 \\
         Standard deviation & 0.074 & 0.074 & 0.074 \\
         \hline
    \end{tabular}
    \caption{Vertical distances [\SI{}{mm}] between the CT mesh and various simplified photogrammetric meshes of the Cuypers sound board on a regular grid (linear interpolation). The original photogrammetric sound board without simplification contains about 420k vertices.}
    \label{tab:Vertical_dist_CT}
\end{table}

\begin{table}[H]
    \centering
    \begin{tabular}{|l|c|c|}
    \hline
          Cuypers sound board & 250k vertices & 50k vertices \\
         \hline
         Maximum distance & $2.8 \cdot 10^{-2}$ & $8.2 \cdot 10^{-1}$ \\
         Mean & $2.0 \cdot 10^{-4}$ & $3.3 \cdot 10^{-3}$ \\
         Median & $3.8 \cdot 10^{-5}$ & $2.5 \cdot 10^{-3}$ \\
         Standard deviation & $3.8 \cdot 10^{-4}$ & $5.3 \cdot 10^{-3}$\\
         \hline
    \end{tabular}
    \caption{Vertical distances [\SI{}{mm}] between the original photogrammetric mesh (about 420k vertices) and two simplified photogrammetric meshes of the Cuypers sound board on a regular grid (linear interpolation).}
    \label{tab:Vertical_dist_photogrammetry}
\end{table}

\section{Geometric analysis of the sound boards} \label{Geometric analysis}
\noindent In this section we highlight several characteristics that can help distinguish a reduced violin from an unreduced instrument, namely the contour lines, the asymmetry between the back and the sound board and the minima channel. Our results are based on the geometric analysis of the photogrammetric meshes obtained and validated in Section \ref{Validation}. A preliminary report on this analysis can be found in \cite{beghindigital}. In order to calculate the three characteristics aforementioned, we need first to compute the plane of symmetry of the violin between the back and the sound board, which will serve as a reference for the computations. 

\subsection{Symmetry plane between the sound board and back}
\label{Symmetry plane between the sound board and back}

\noindent We explained in Section \ref{Contour isolation} that we oriented the body of the violin with a PCA before delineating the contours of the sound board and back. However, as the point cloud of the body contains the ribs and some artefacts, the plane of the PCA does not exactly match what can be considered the natural horizontal plane of symmetry between the sound board and back, although they are close. We therefore propose a way to correct the orientation of this symmetry plane, which is crucial because we will use it to calculate the contour lines, quantify the asymmetry and identify the channel of minima. This reorientation does not put us at odds with the validation performed in Section \ref{Validation} as we are only applying a rotation operator to our mesh. Thus, the optimised angles will differ slightly from those in Table \ref{tab:MinProb} but the overall result leads to the same average distance $D$. Furthermore, as the orientation of this symmetry plane is very close to that of the plane originally identified by the PCA, we consider that the contour isolation proposed in Section \ref{Contour isolation} (which depended on the PCA orientation) is still coherent and entirely suitable for our analysis. \\

\noindent We need however to mention clearly that the `plane of symmetry' of a violin is a misnomer. There is no real plane between the sound board and the back. First, the ribs are generally smaller near the end of the sound board and back than on the rest of the body. Also, wood ages and warps over time. What we defined here as a plane of symmetry is the closest notion to an ideal symmetry and best conceptualises something that does not actually exist. \\

\noindent We identify this plane of symmetry using the individual orientations of both the sound board and the back, by calculating the best plane that passes through each of the two surfaces and `averaging' them (i.e. the average plane of symmetry is the plane bisecting the acute angle between the planes of the sound board and back). We then rotate the meshes to make this average plane of symmetry parallel to the horizontal plane $\Pi\equiv z=0$ and we finally adjust its offset (see later in this section).\\

\begin{figure}[H]
    \centering
    \includegraphics[scale = 0.5]{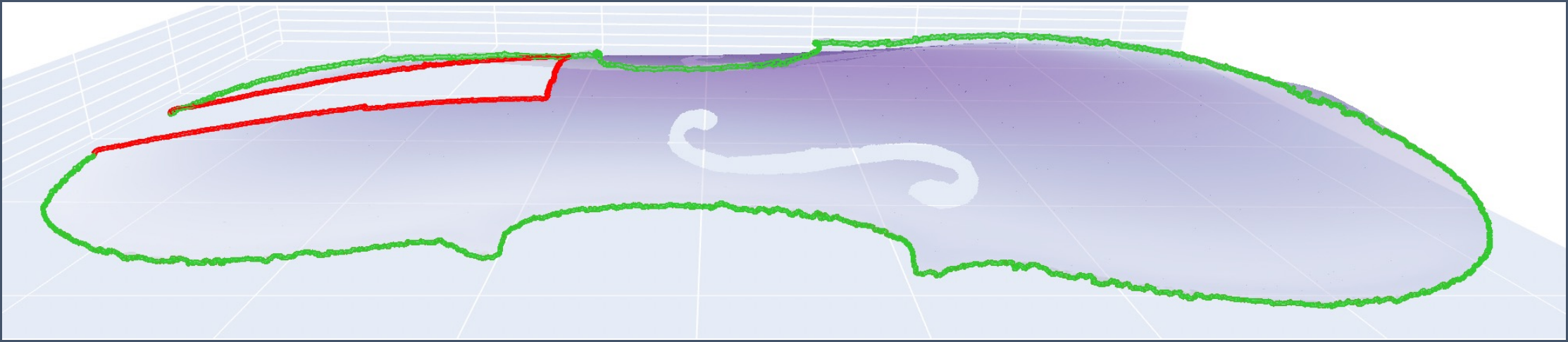}
    \caption{Contour of the sound board (green) with highlight on the raised part to be manually removed (red).}
    \label{fig:contour_manual}
\end{figure}

\noindent We compute each of the best planes approximating the sound board and back with an orthogonal regression, which does not favour any direction (and removes any influence from the initial axes computed by PCA). We actually consider three\footnote{We also tried to apply an orthogonal regression on the whole cloud (both the sound board and back as a single cloud), but it did not lead to convincing results. Indeed, because of the different number of vertices between the two surfaces, the regression is biased towards the larger pointcloud.} options for this procedure before averaging those planes. The orthogonal regressions are therefore performed:
 
\begin{itemize}
    \item on all the vertices from the sound board or the back, considered as two independent meshes (`Two meshes').  
    \item on the vertices of the contour of the sound board or the back (as computed in Section \ref{Contour isolation}) (`Two contours').
    \item on the vertices of the contour of the sound board or the back (as computed in Section \ref{Contour isolation}), with the raised part of the sound board removed manually (`Two contours (manual)'). Indeed, because of our isolation process, the contour of the sound board contains a raised part that biases the regression (see Figure \ref{fig:contour_manual}).
\end{itemize}

\noindent For each of these three configurations, we calculated the angle between the average plane of symmetry (before rotation) and the horizontal plane $\Pi\equiv z=0$ for Hofmans' instrument. The results are given in Table \ref{tab:orthogonal_regression}. All three angles are very similar and indicate that a realignment of the violins was necessary. Because of their close value, we finally retained the configuration that made the most sense to us, namely the third option `Two contours (manual)'. Indeed, the wooden board used by the luthiers to build the sound board and the back is flat on its inside. Thus, the manually corrected contours best characterise what we mean by plane of symmetry and `horizontality'. Moreover, the angle provided by this approach is almost the average between the other two. For information, the values in Table \ref{tab:orthogonal_regression} are nearly identical with a linear regression (least squares) rather than an orthogonal regression.

\begin{table}[H]
    \centering
    \begin{tabular}{|l|c|}
    \hline
    Configuration & Angle [$\degree$]  \\
         \hline
         Two meshes & $1.62$  \\
         Two contours & $1.15$   \\
         Two contours (manual) & $1.40$ \\
         \hline
    \end{tabular}
    \caption{Orthogonal regression on three configurations and angle between the average symmetry plane (before rotation) and the horizontal plane for Hofmans' instrument.}
    \label{tab:orthogonal_regression}
\end{table}

\begin{figure}[H]
    \centering
    \includegraphics[height = 4 cm]{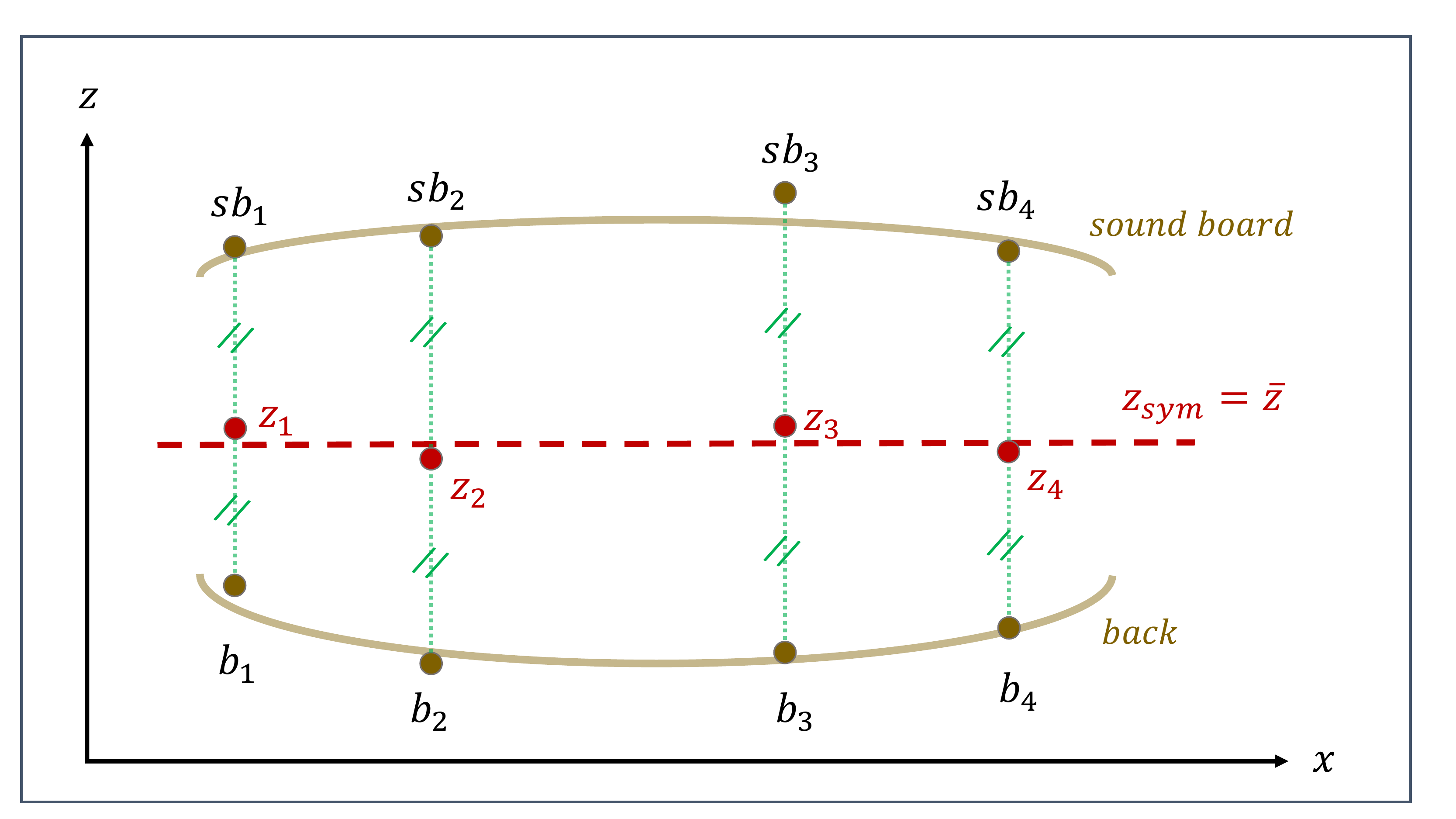}
    \caption{Computation of the offset of the plane of symmetry.}
    \label{fig:z_sym}
\end{figure}

\noindent We apply now a rotation so that the average plane of symmetry is horizontal and we adjust its offset. To do so, we compute the $z-$values of the sound board and the back on a horizontal regular grid (after rotation) with nodes equally spaced every $\SI{1}{mm} \times \SI{1}{mm}$. As in Section \ref{Simplification}, we draw a vertical normal for each node $i$ of the grid, and identify the points through which this normal intersects the surfaces. We denote these intersection points $sb_i$ for the sound board and $b_i$ for the back respectively. We then compute, for each node $i$ of the grid, the mean point $z_i = \frac{sb_i+b_i}{2}$ located at equal distance from the point $sb_i$ of the sound board and its corresponding point $b_i$ on the back. If one of the two points $sb_i$ or $b_i$ is not defined on a node $i$ of the grid, we do not calculate $z_i$ at this node (this happens for example for sound holes, that are empty on the sound board but not on the back). We then calculate the offset of the horizontal plane by averaging all midpoints, $z_{sym} = \bar{z} =  \frac{1}{N_g} \sum_{i=1}^{N_g} z_i$, where $N_g$ is the total number of valid nodes on the grid, i.e. for which $z_i$ is defined. Finally, now that the offset is calculated, we shift the meshes so that the symmetry plane matches the plane $\Pi\equiv z=0$. The points of the shifted sound board and back are denoted $sb_{i,shift} = sb_i - \bar{z}$ and $b_{i,shift} = b_i - \bar{z}$. Figure \ref{fig:z_sym} shows a 2D example of the calculation of the offset of the plane of symmetry $z_{sym}$ before the shift.  

\subsection{Contour lines}
\label{Contour lines}
\begin{figure}[H]
\centering
\begin{subfigure}[b]{0.475\textwidth}
\includegraphics[scale = 0.4]{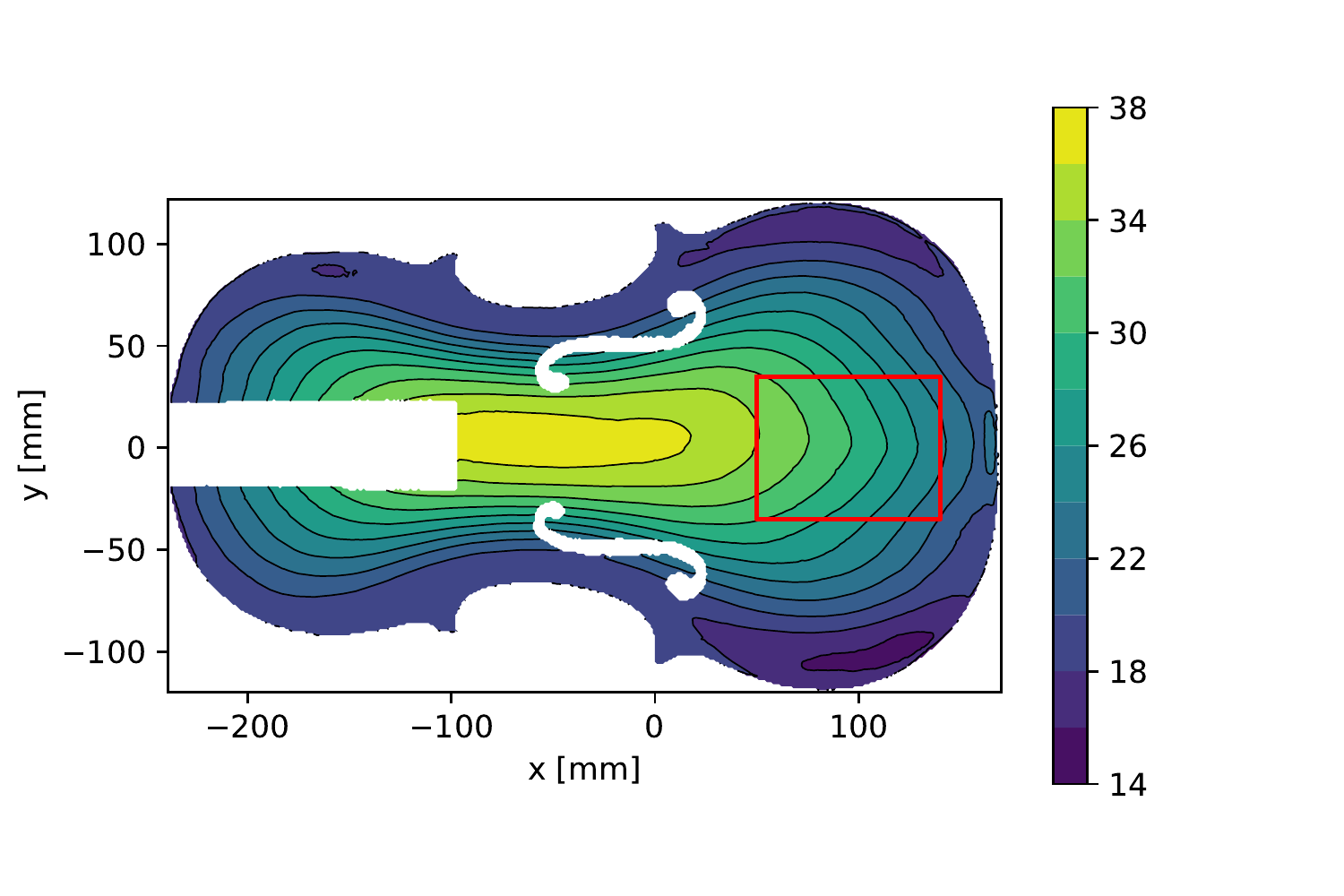}
\vspace{\baselineskip}
\includegraphics[scale = 0.4]{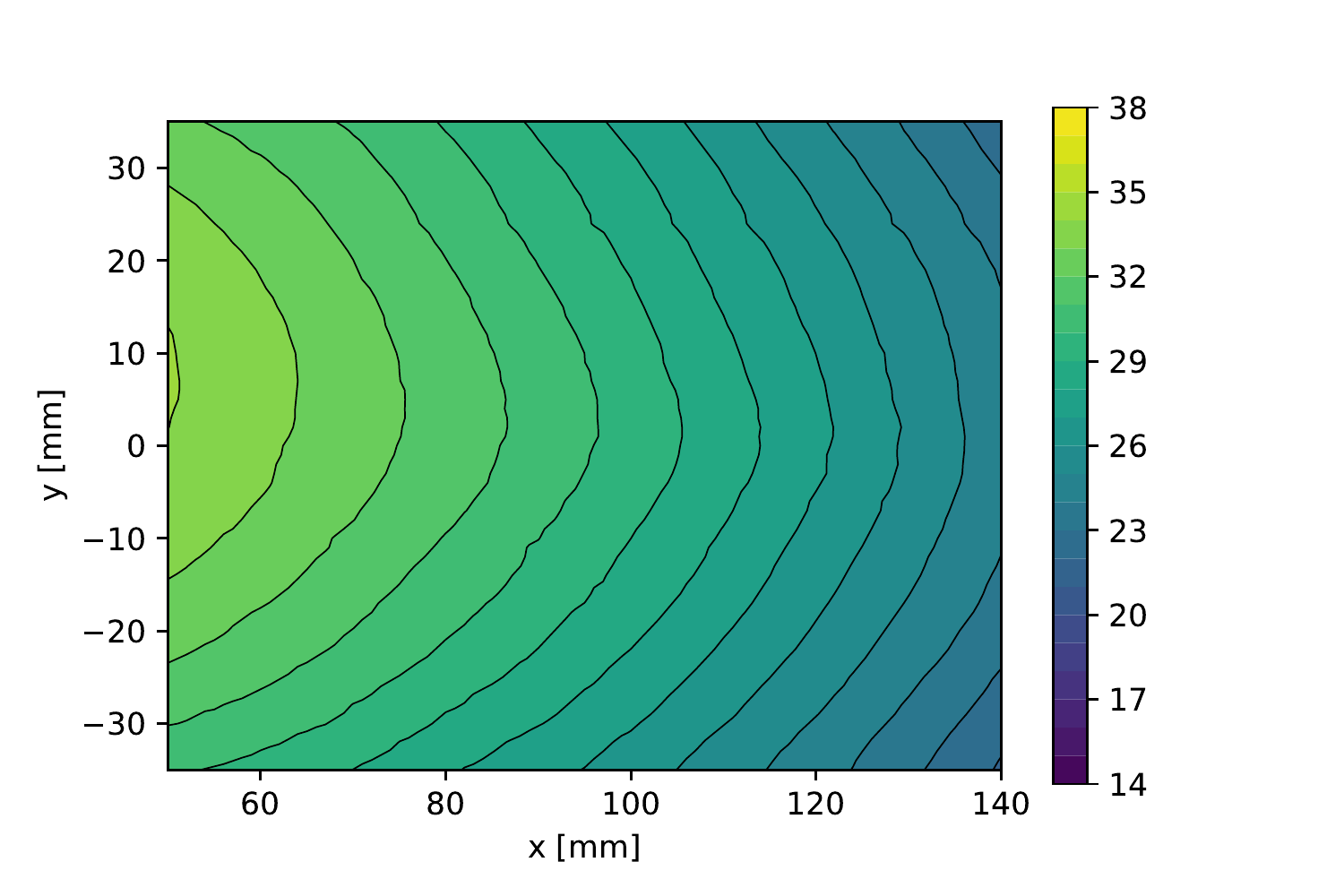}
\end{subfigure}%
\begin{subfigure}[b]{0.475\textwidth}
\includegraphics[scale = 0.4]{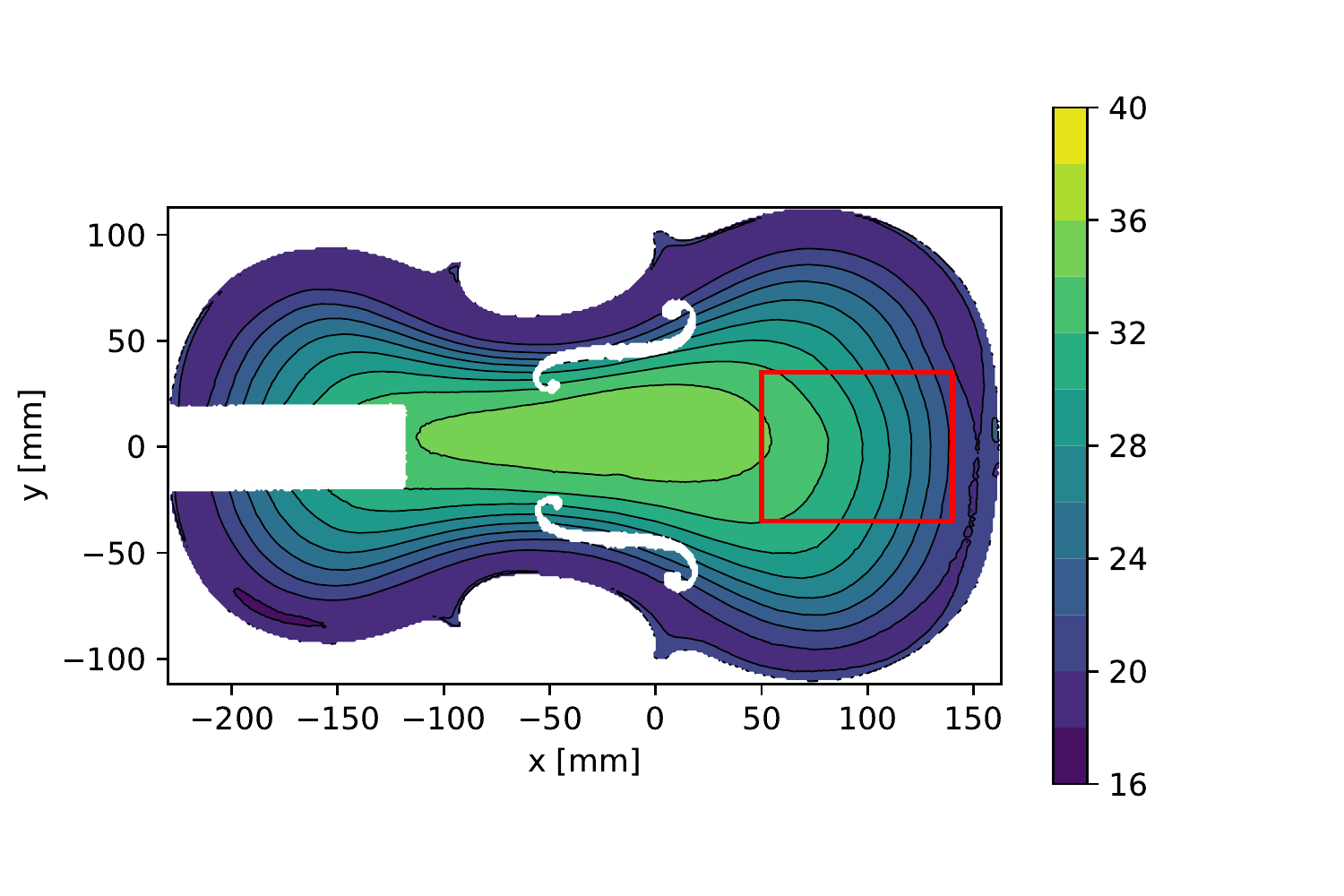}
\vspace{\baselineskip}
\includegraphics[scale = 0.4]{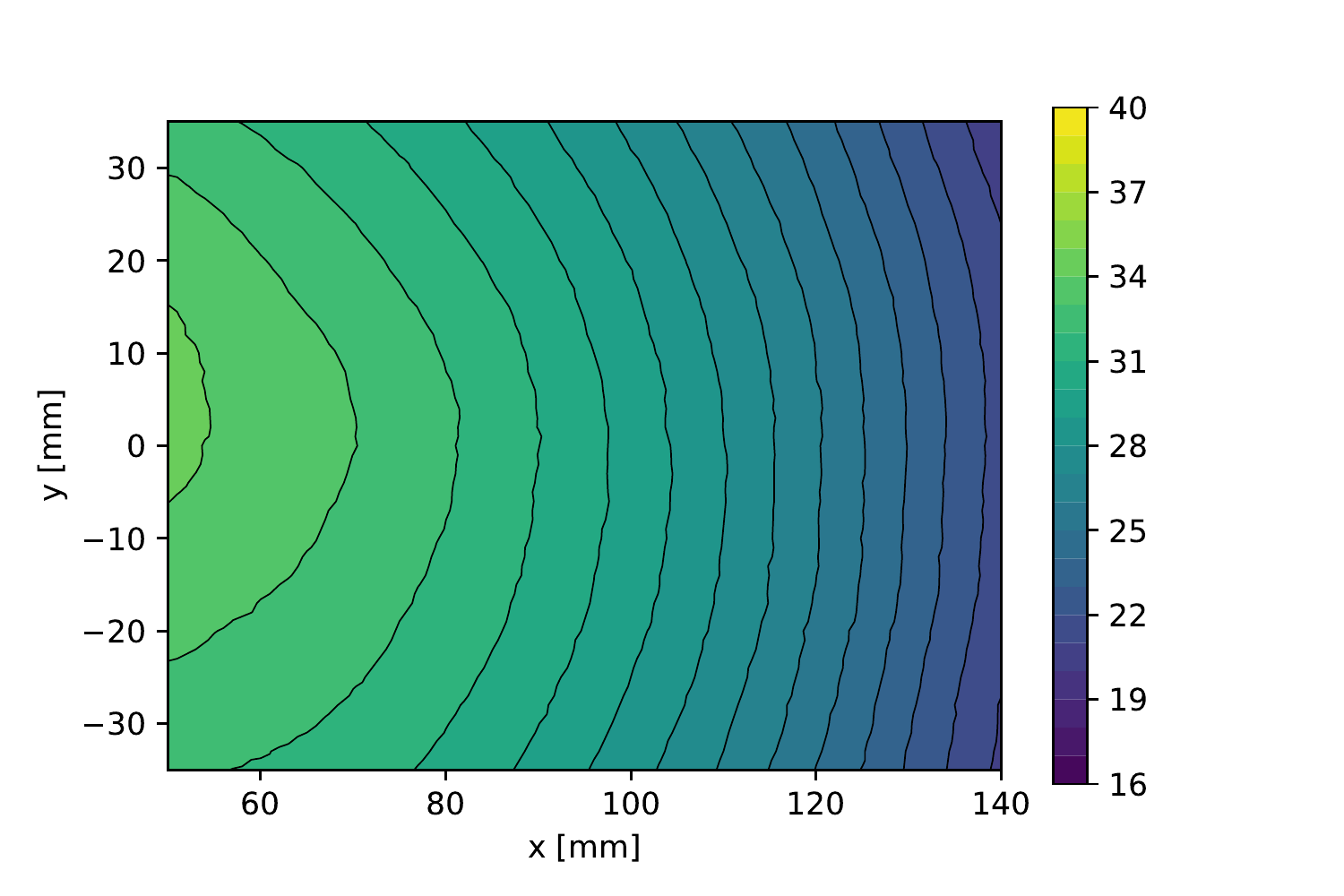}
\end{subfigure}%
\caption{Contour lines of the Hofmans (left) and Cuypers (right) sound boards [\SI{}{\mm}].}
\label{fig:contour_lines_sb}
\end{figure}

\begin{figure}[H]
\centering
\begin{subfigure}[b]{0.475\textwidth}
\includegraphics[scale = 0.4]{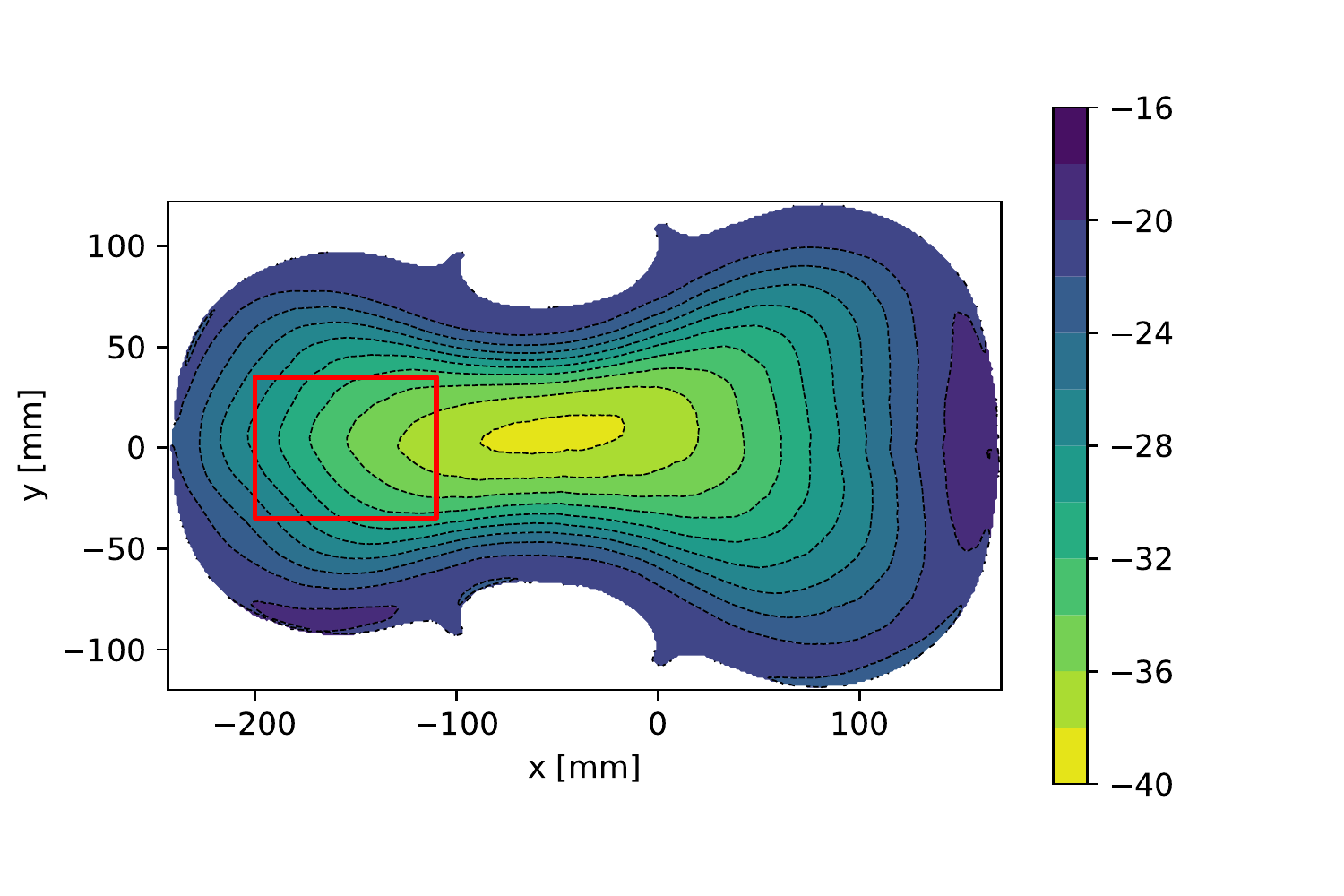}
\vspace{\baselineskip}
\includegraphics[scale = 0.4]{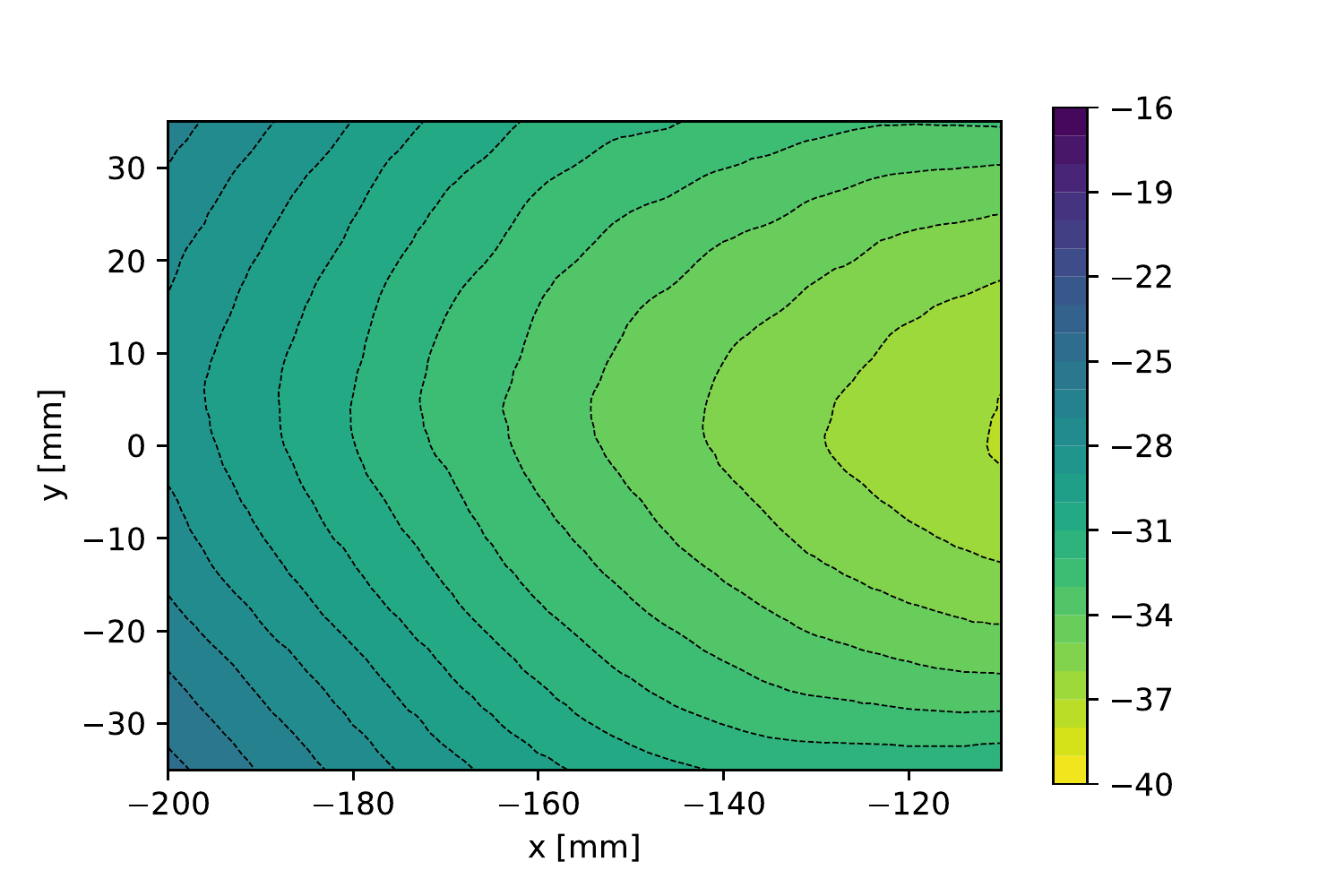}
\end{subfigure}%
\begin{subfigure}[b]{0.475\textwidth}
\includegraphics[scale = 0.4]{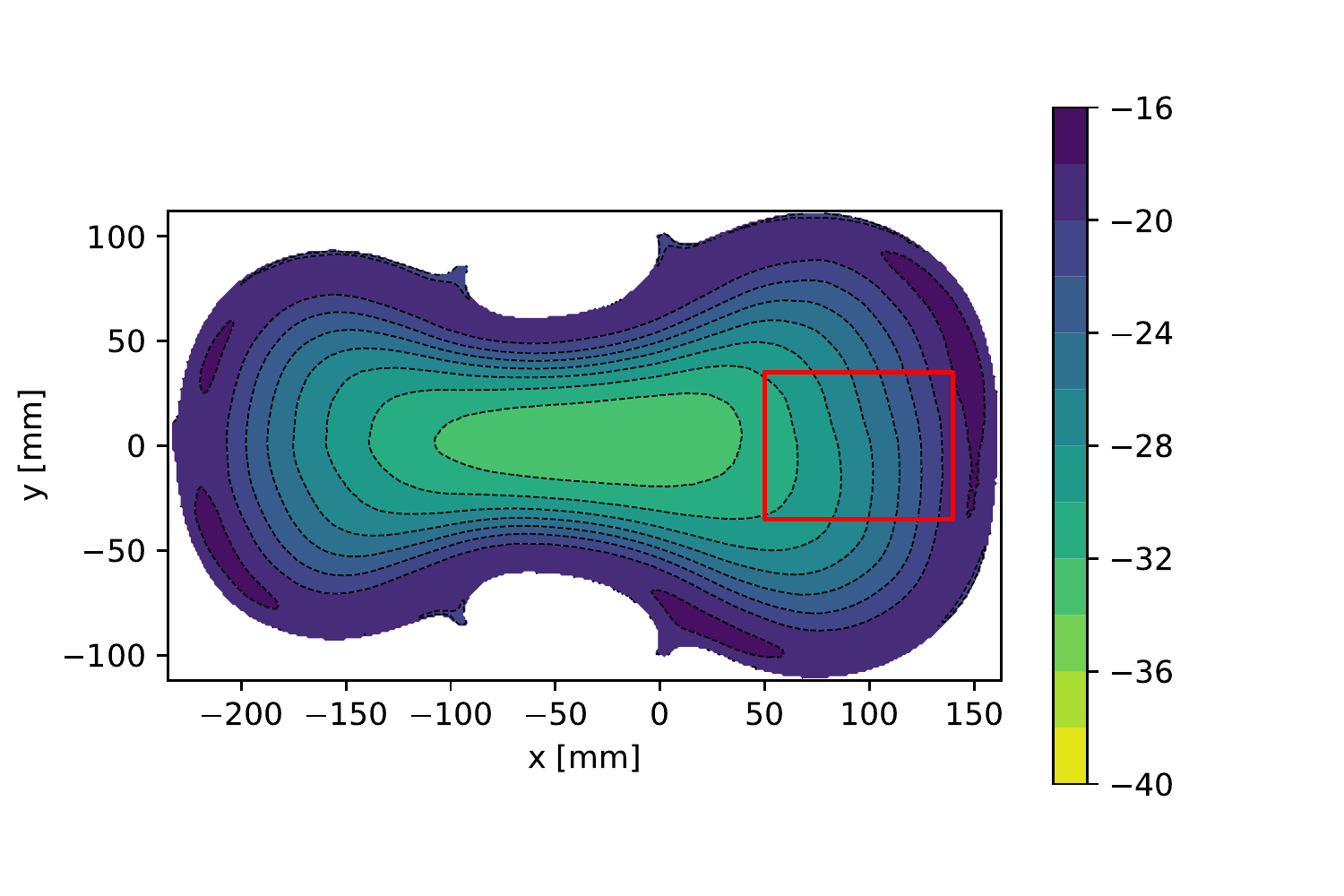}
\vspace{\baselineskip}
\includegraphics[scale = 0.4]{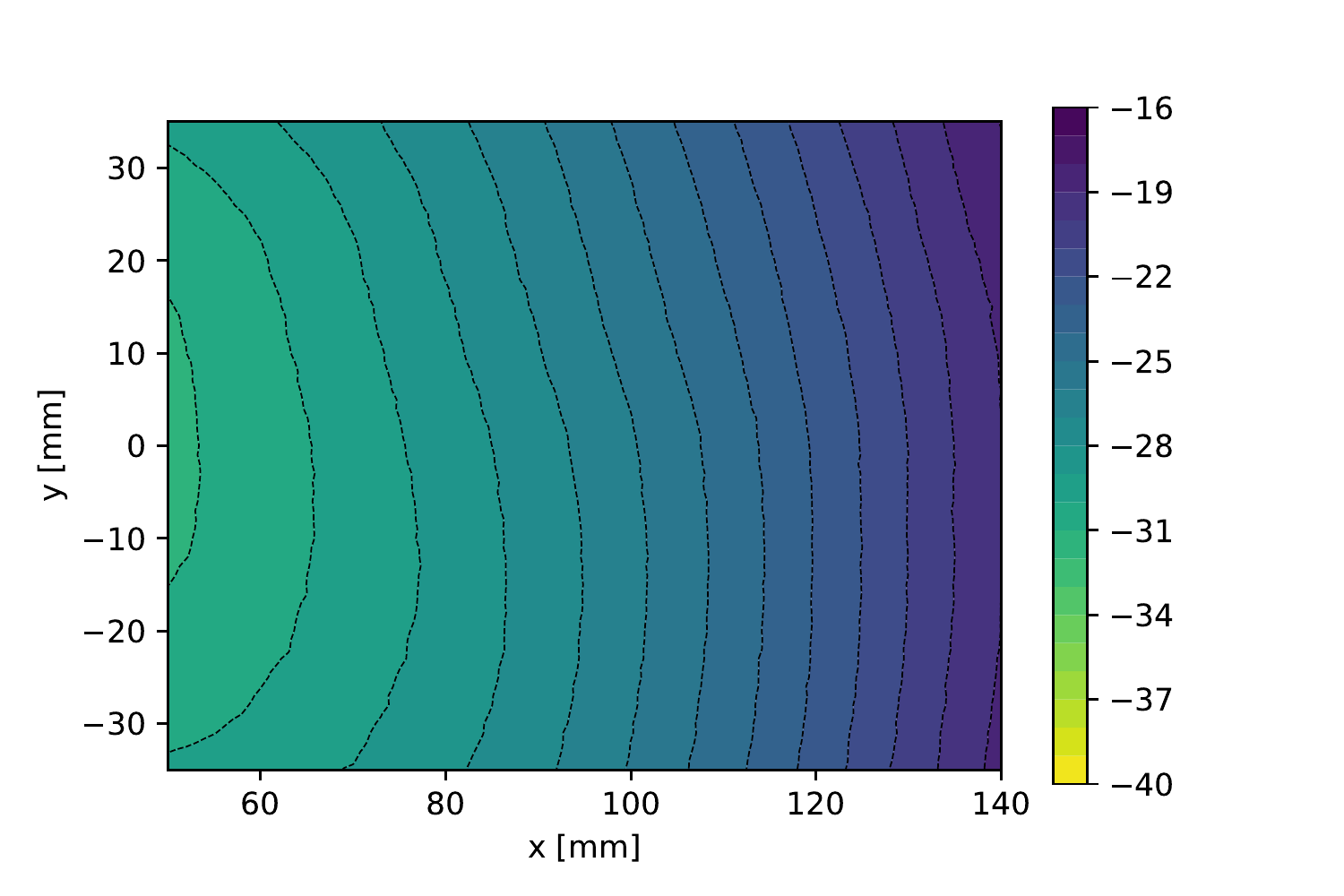}
\end{subfigure}%
\caption{Contour lines of the Hofmans (left) and Cuypers (right) backs [\SI{}{\mm}].}
\label{fig:contour_lines_back}
\end{figure}

\noindent We compute horizontal sections of four surfaces (two sound boards and two backs) every \SI{2}{\mm} based on the symmetry plane defined in Section \ref{Symmetry plane between the sound board and back}. The four sets of contour lines are represented according to the same relative convention: the level closest to the plane of symmetry is in dark blue and the range of the altitude is up to \SI{24}{\mm} from this closest level. The sound board is to be seen as a `hill' while the back is to be seen as a `valley'. In addition, positive contour lines (sound board) are represented with continuous lines and negative lines (back) are dashed. Figure \ref{fig:contour_lines_sb} shows, especially in the zoomed area (red frame, refinement every \SI{}{\mm}), that the contour lines are rounder on the unreduced Cuypers, and sharper on the Hofmans. We suppose that this sharpness is due to a slice of wood removed along the main axis of the violin (width reduction), as illustrated in Figure \ref{fig:Moens} (right). A similar behaviour is also observed for the back of both instruments in Figure \ref{fig:contour_lines_back}. Finally, it is worth noting that the contour lines at the bottom of the Hofmans back are almost perpendicular to the main axis of the instrument. 

\subsection{Asymmetry between sound board and back}
\label{Asymmetry}

\noindent Interestingly, when a violin is reduced, the sound board and back do not necessarily follow the same reduction pattern. Hence we are interested in studying the asymmetry between the two surfaces facing each other. To do so, we compute the vertical differences between the sound board and the back on a horizontal regular grid with nodes equally spaced every $\SI{1}{mm} \times \SI{1}{mm}$. We re-use the values $sb_{i,shift}$ and $b_{i,shift}$ (or equivalently $sb_i$, $b_i$, $z_i$ and $\bar{z}$) from Section \ref{Symmetry plane between the sound board and back} and we calculate the asymmetry $a_i$ at any point on the grid as the difference in the distances of the sound board and the back from the horizontal plane, which is a signed quantity: 
\begin{equation}
\begin{split}
a_i &= sb_{i,shift} - |b_{i,shift}| \\
&= (sb_i - \bar{z}) - |b_i-\bar{z}| \\
    &= 2(z_i-\bar{z})
\end{split}
\tag{A}
\label{eq:A}
\end{equation}
assuming that $sb_{i,shift} > 0$ and $b_{i,shift} < 0$ (or equivalently, $sb_i > \bar{z}$ and $b_i < \bar{z}$). The asymmetry $a_i$ is not defined at nodes $i$ for which either $sb_i$ or $b_i$ is not defined.  \\

\begin{figure}[H]
\centering
\begin{subfigure}[b]{0.475\textwidth}
\includegraphics[scale = 0.45]{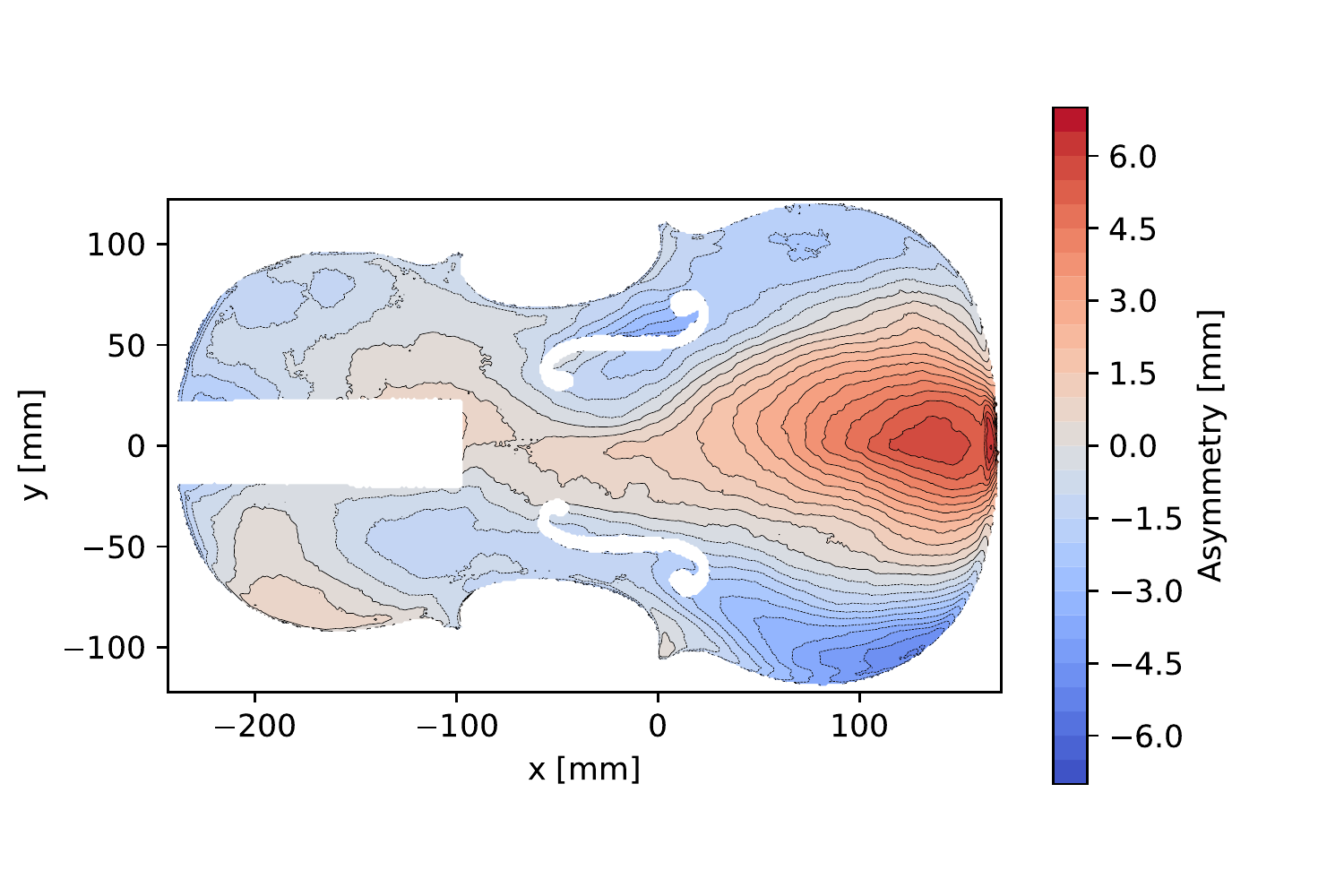}
\vspace{\baselineskip}
\includegraphics[scale = 0.4]{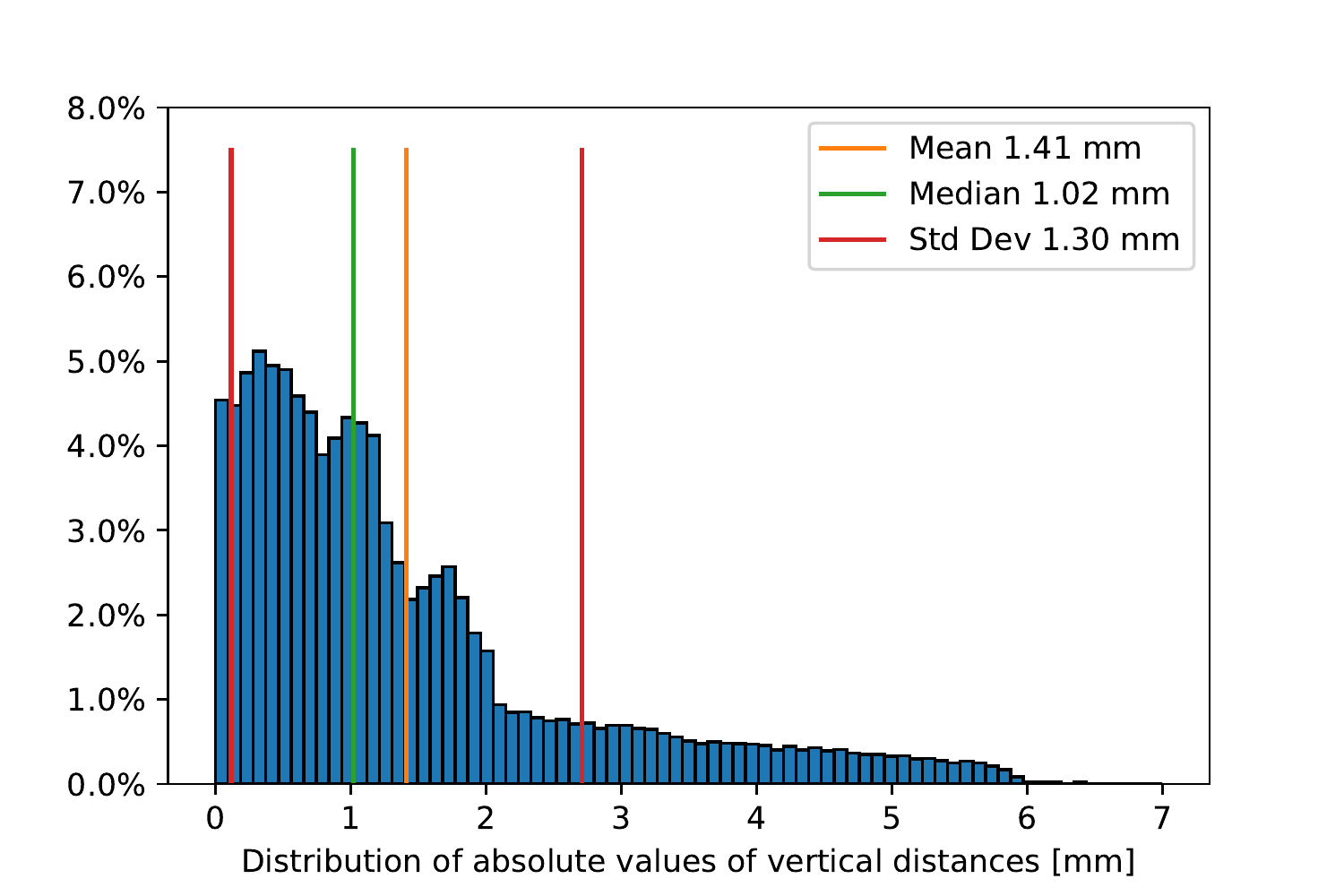}
\end{subfigure}%
\begin{subfigure}[b]{0.475\textwidth}
\includegraphics[scale = 0.45]{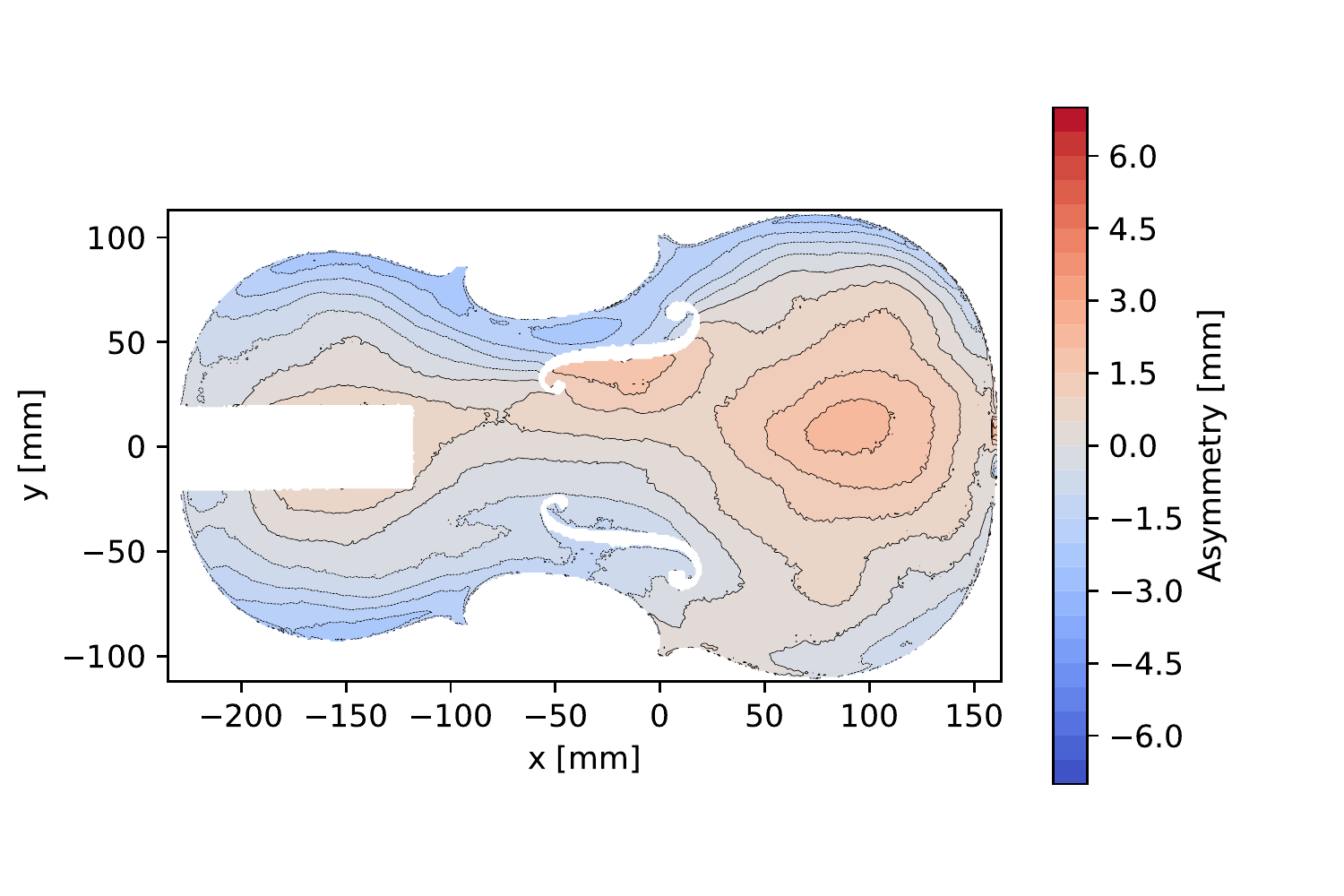}
\vspace{\baselineskip}
\includegraphics[scale = 0.4]{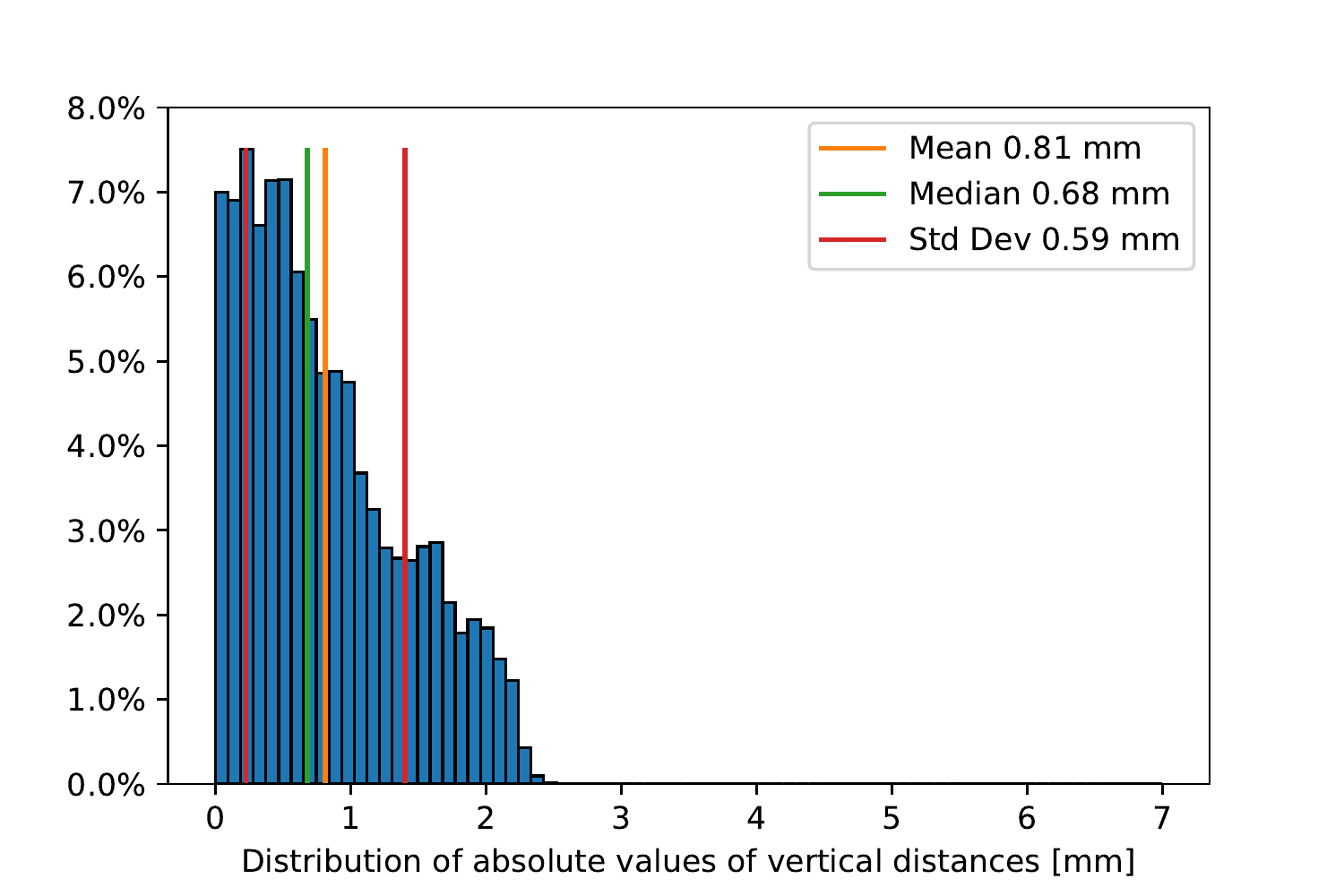}
\end{subfigure}%
\caption{Heat map of the asymmetry between the sound board and the back (top) and distribution of the absolute values of the vertical distances (bottom) for the Hofmans (left) and Cuypers (right). For the heat maps, a positive value means that the sound board is further from the plane of symmetry than the back and a negative value means that the back is further from the plane of symmetry than the sound board.}
\label{fig:symmetry}
\end{figure}

\noindent Figure \ref{fig:symmetry} shows the topography of the vertical distances and their (absolute) distribution. A positive value means that the sound board is further from the plane of symmetry than the back and a negative value means that the back is
further from the plane of symmetry than the sound board. We also provide histograms of the distribution of the absolute values of all distances for each instrument.

\noindent We immediately see that the difference between the sound board and the back of the Hofmans instrument is much more pronounced than the one of the Cuypers. The distances go up to almost \SI{7}{\mm} for the reduced violin while they stop at \SI{2.5}{\mm} for the unreduced one. \\

\noindent Analysing two instruments is not sufficient to draw general conclusions, but we believe that this technique provides interesting and relevant insights about the presence of a reduced violin. 

\subsection{Channel of minima}
\label{Channel of minima}

\begin{figure} [H]
\begin{subfigure}{0.48\textwidth}
\centering
\includegraphics[scale = 0.42]{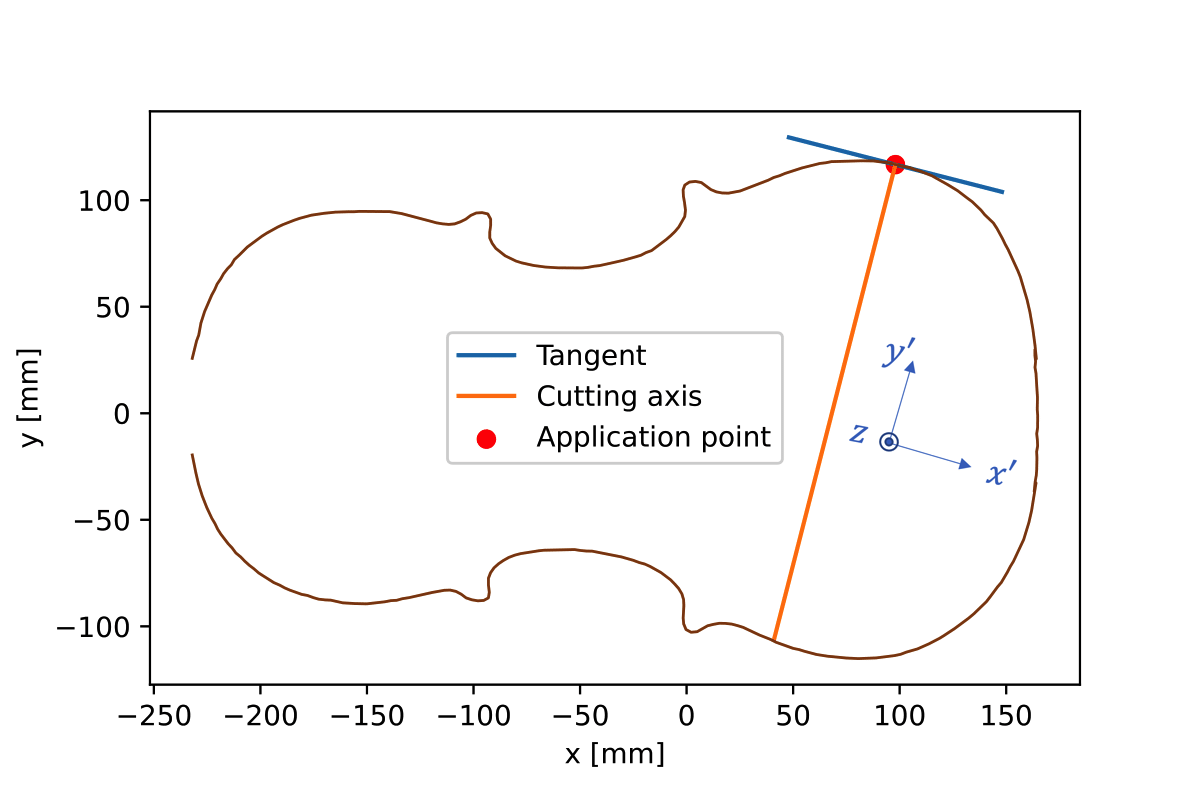}
\end{subfigure}
\begin{subfigure}{0.48\textwidth}
\centering
\includegraphics[scale = 0.26]{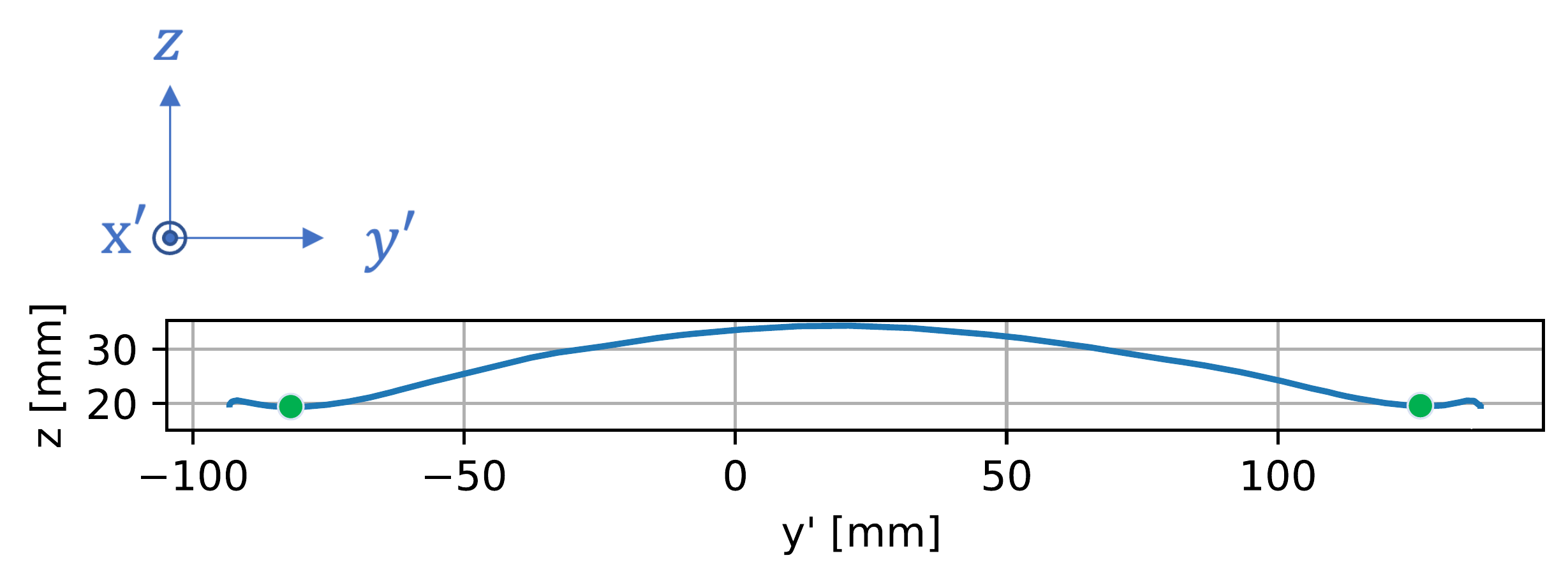}
\end{subfigure}
\caption{Top view (left) and cross-section (right) of the Hofmans sound board. The interpolation of the contour with cubic splines is in brown in the left figure.}
\label{fig:cut2D}
\end{figure}

\noindent The sound board and the back of a violin feature a `channel of minima' running close to their outer contour. To identify this channel, we first interpolate the vertices of the contour of the sound board or the back (obtained with the procedure in Section \ref{Contour isolation} and reoriented in Section \ref{Symmetry plane between the sound board and back}) using cubic splines. Then, we compute a large number of cross-sections through the mesh (separately for the sound board and back). These sections are orthogonal to the symmetry plane from Section \ref{Symmetry plane between the sound board and back} and chosen to be perpendicular to the tangents of the sound board contour (computed from the cubic spline interpolation), as shown in orange in Figure \ref{fig:cut2D} (left). In each cross-section, minima are identified as the points with the lowest $z-$height among those close to the tangent point. They can be seen in green in Figure \ref{fig:cut2D} (right). The channel of minima of the backs shows a similar behaviour.  

\begin{figure}[H]
\centering
\begin{subfigure}[b]{0.475\textwidth}
\includegraphics[scale = 0.4]{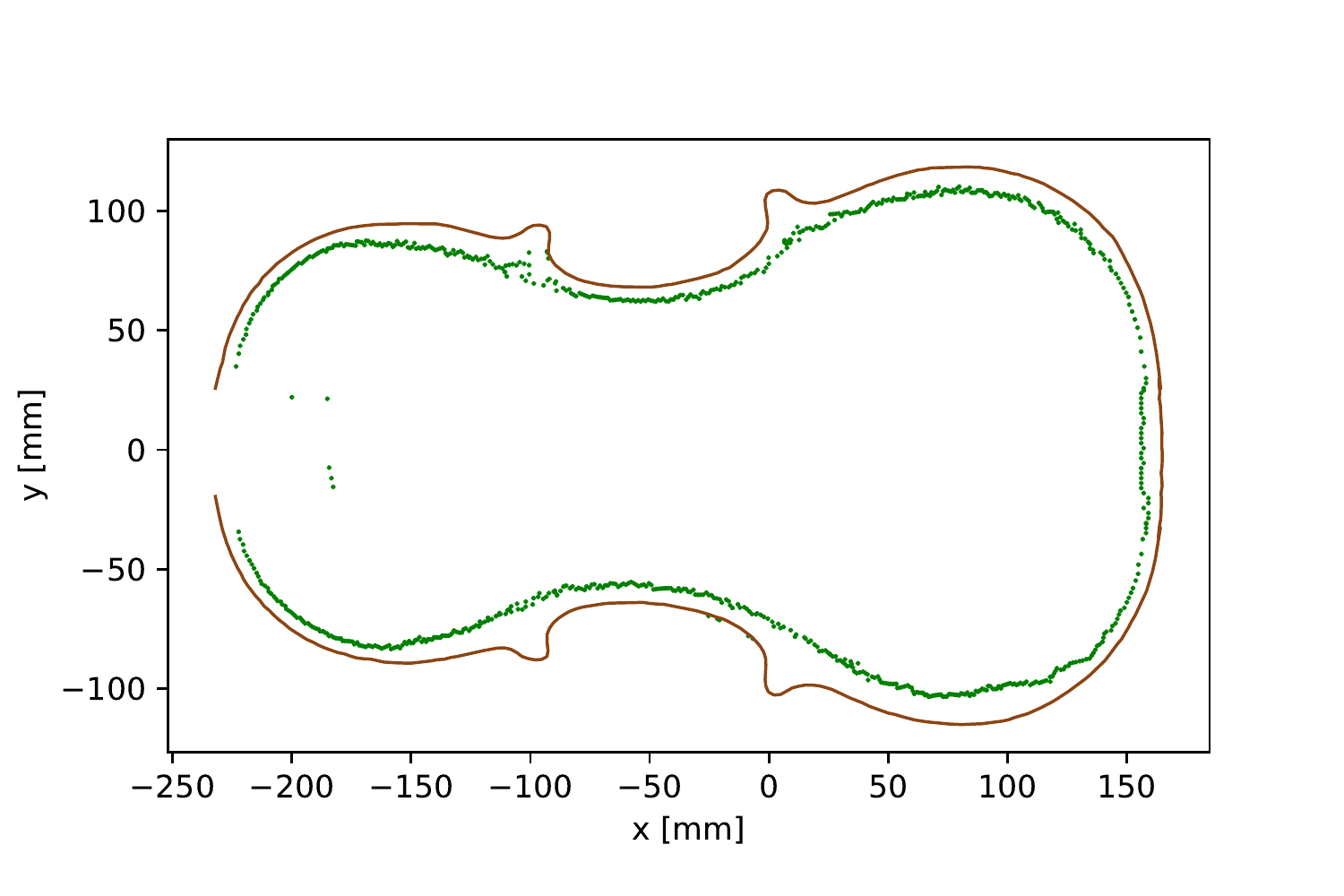}
\vspace{\baselineskip}
\includegraphics[scale = 0.4]{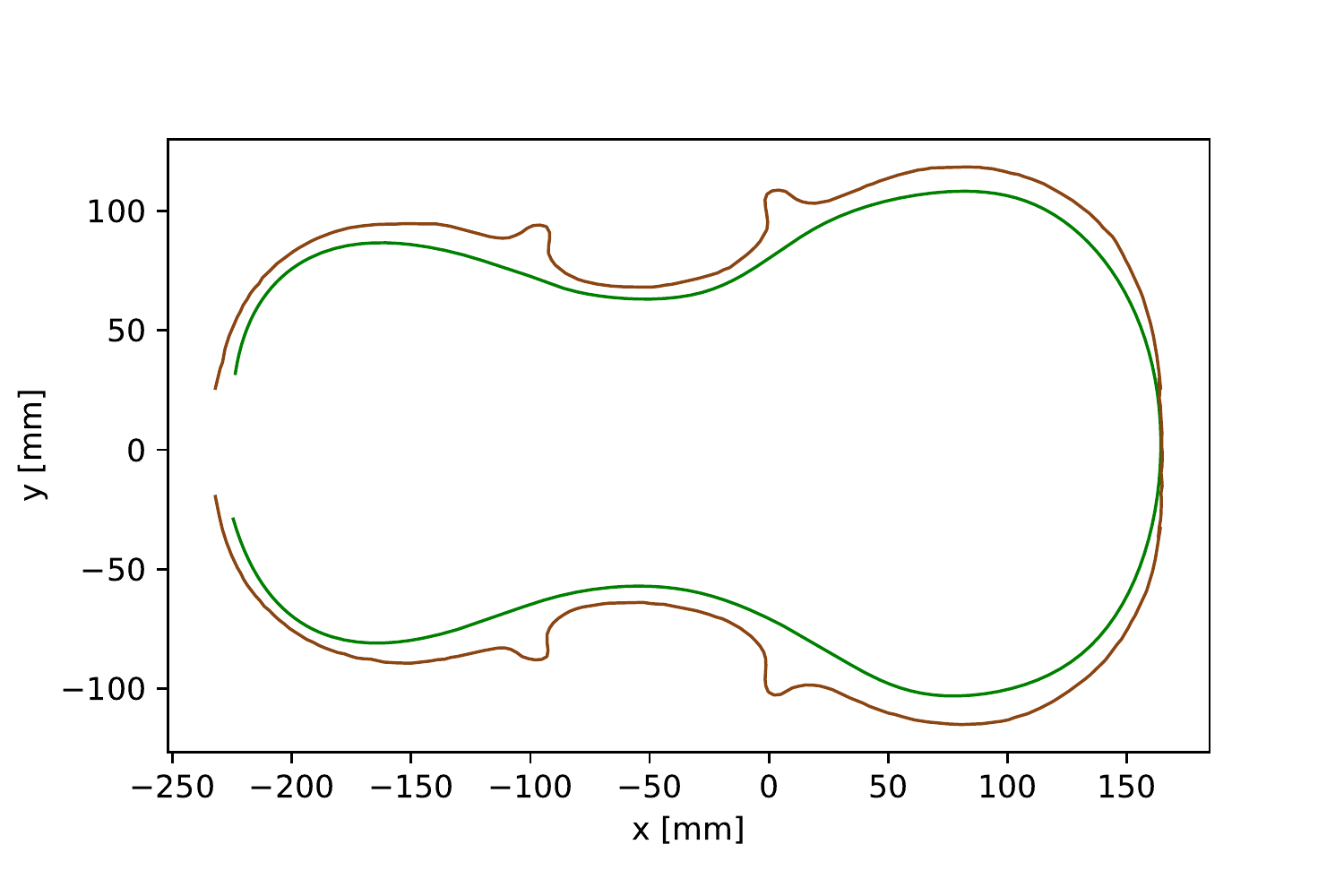}
\end{subfigure}
\begin{subfigure}[b]{0.475\textwidth}
\includegraphics[scale = 0.4]{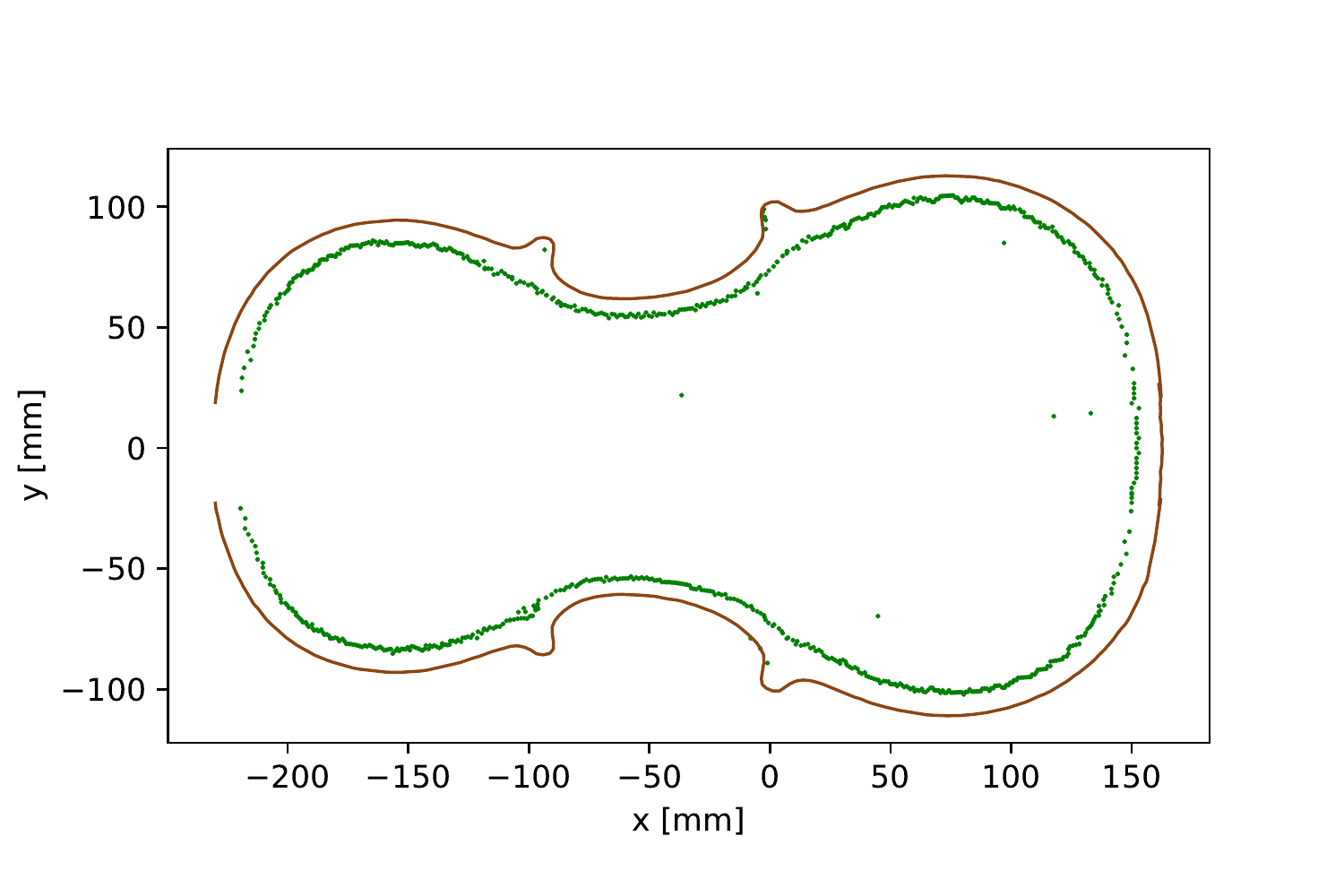}
\vspace{\baselineskip}
\includegraphics[scale = 0.4]{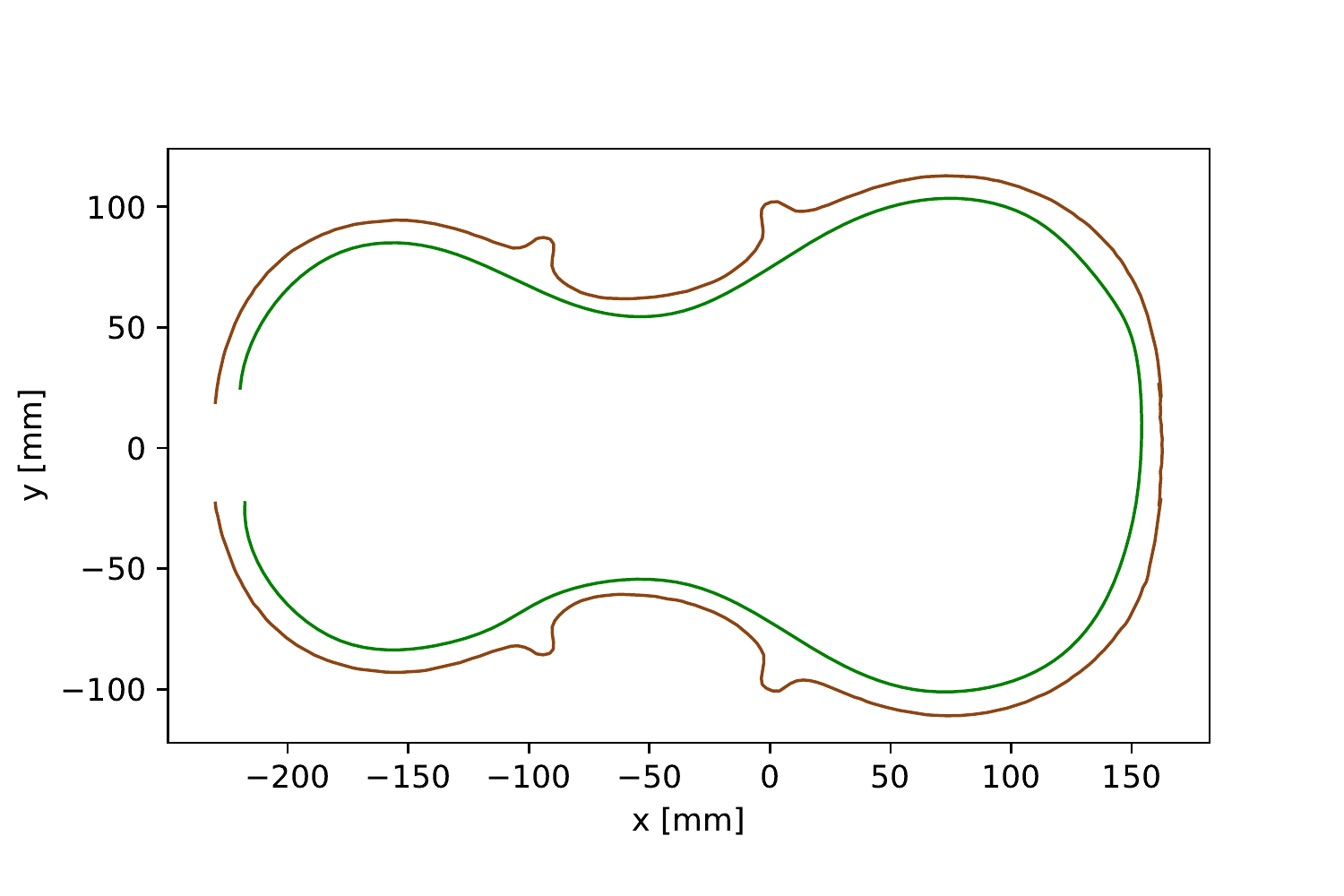}
\end{subfigure}%
\caption{Channel of minima for the Hofmans (left) and Cuypers (right) sound boards. \\
Raw data (top) and spline approximation (bottom).}
\label{fig:channels}
\end{figure}

\noindent The `raw' channel of minima is shown in Figure \ref{fig:channels} (top), exhibiting clear differences between the instruments. Indeed, we see that, in a reduced violin, the distance from the channel to the contour tends to decrease in some areas close to the top and the bottom of the sound board. Note however that the apparent recess in the channel at the bottom of the Hofmans sound board is due to the lower nut (the small ebony rim over which the strings pass), and not to the actual channel, which has disappeared at this point (see \cite{GPS} for a detailed explanation). The spline approximation of the channel displayed in Figure \ref{fig:channels} (bottom) shows a more realistic trace. We finally mention that a similar behaviour is also observable for the back surfaces.

\section{Conclusion and future work}
\label{Conclusion}

\noindent We proposed a geometric approach for the objective study of early violins based on photogrammetric meshes, validated with sub-millimetre accuracy by comparison to reference CT scans. The accuracy of photogrammetry is similar to that obtained by \cite{pinto2008photogrammetric} for violin reconstruction, but has been validated with a physical representation of the instrument, and not with a synthetic version of it. Our accuracy is also similar to that obtained in \cite{ho2017comparing, donato2020photogrammetry}, also comparing photogrammetry to CT scans (of human skulls), but we validate it here with geometric and not statistical tools. \\

\noindent After confirming the validity of our photogrammetric approach, we showed how the computation of morphological characteristics such as contour lines, asymmetry and channel of minima allows to characterise whether or not instruments were reduced. To the best of our knowledge, this type of objective approach, based on the geometric analysis of three-dimensional meshes, has not been considered yet in the literature. As the comparison was made for only two instruments, it would be somewhat risky to attempt to generalise our results immediately to all violins, whether or not they were reduced. However, the conclusions are encouraging. In the future, we plan to apply our techniques to a larger collection of approximately forty instruments including violins, violas and cellos. We hope that this corpus will allow us to detect automatically features of reduced instruments, using clustering or classification techniques applied to appropriate mathematical representations of the surface of the sound boards and backs. \\

\noindent Despite the fact that we start from three-dimensional data (meshes and point clouds), some parts of our geometric analysis rely on two-dimensional techniques (contour isolation, use of cross-sections). Developing an exclusively three-dimensional processing is left for future research, and would allow us to consider other features such as the location of the inflection points, and identifying the true shape of the minima channel as a three-dimensional curve. Our ultimate goal would be to predict what the original dimensions of the reduced instruments were, by quantifying the removed crescent of wood at the top and bottom of the sound box and/or the slice of wood along the axis, and comparing them with unreduced violins.

 \section*{Acknowledgements}
 \noindent The authors thank Iona Thys (Royal Museums of Art and History, Brussels) for her photogrammetric work, and Alain  Vlassenbroek, Emmanuel Coche, Etienne Danse and the University Hospital Saint-Luc (UCLouvain, Brussels-Woluwe) for their help with violin CT scans. We also thank Jean-Philippe Echard (Musée de la Musique, Paris) for the interesting discussions we shared. This research did not receive any specific grant from funding agencies in the public, commercial, or not-for-profit sectors. 


\appendix

\section{Comment on the error assessment and validation}
\label{Appendix}

\noindent In Section \ref{Error assessment and validation} we compared CT and photogrammetric point clouds from two instruments. We obtained mean errors of \SI{0.301}{\mm} and \SI{0.215}{\mm} (point-to-point metric) and we saw that the heat maps showed an excellent agreement between the representations. For the sake of experiment and to better interpret those average errors, we made two additional comparisons for crossed instruments, i.e. CT Cuypers vs. photogrammetric Hofmans and CT Hofmans vs. photogrammetric Cuypers. 
\begin{figure}[H]
\centering
\vspace{-0.45 cm}
\begin{subfigure}[b]{0.48\textwidth}
\includegraphics[height = 4.6 cm]{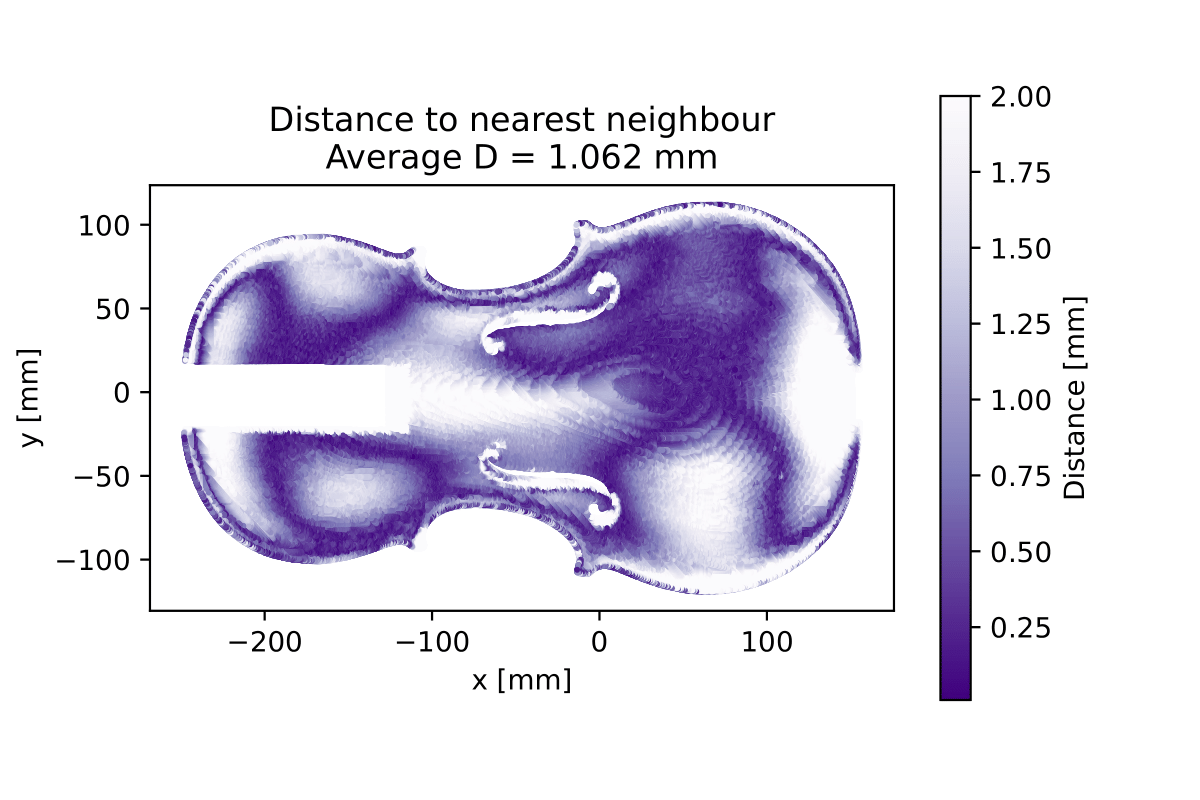}
\vspace{0.01\baselineskip}
\includegraphics[height = 3.9 cm]{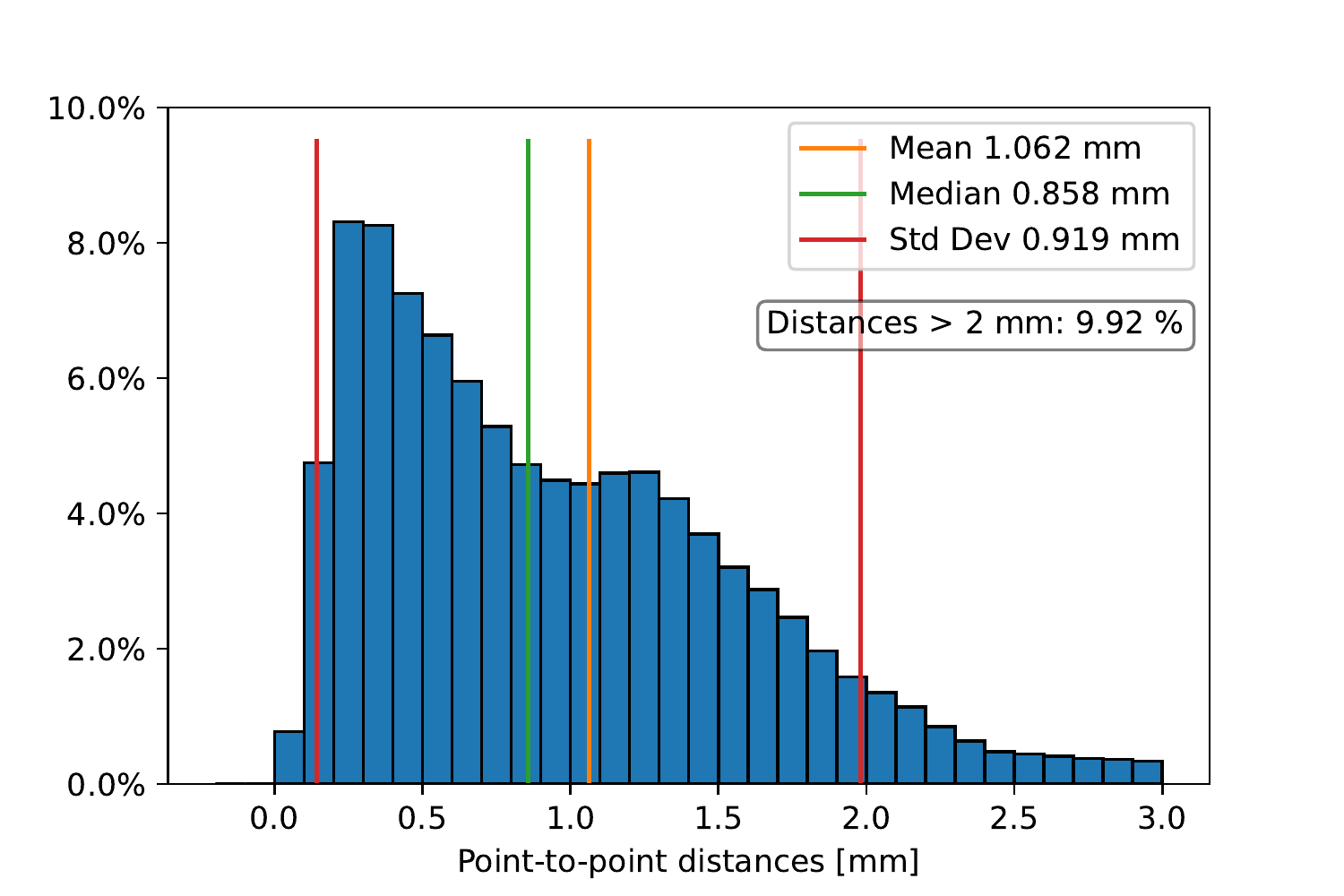}
\end{subfigure}%
\begin{subfigure}[b]{0.48\textwidth}
\includegraphics[height = 4.6 cm]{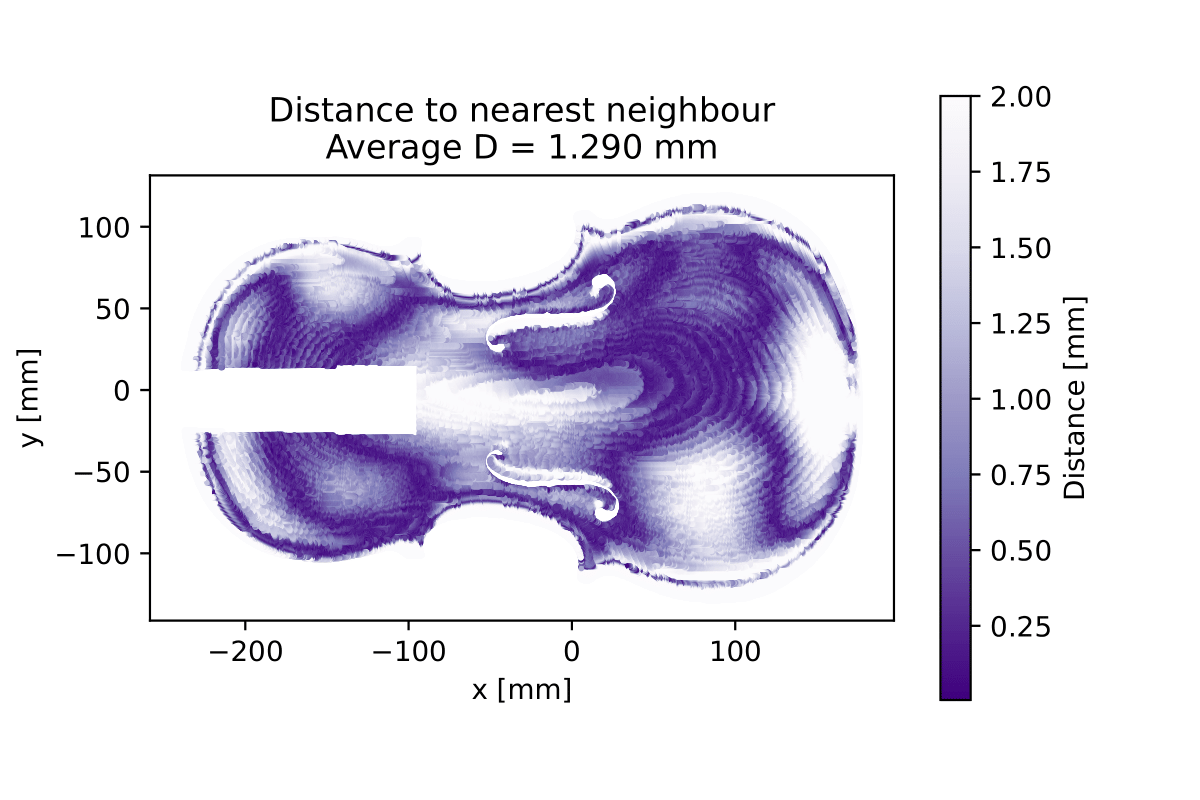}
\vspace{0.01\baselineskip}
\includegraphics[height = 3.9 cm]{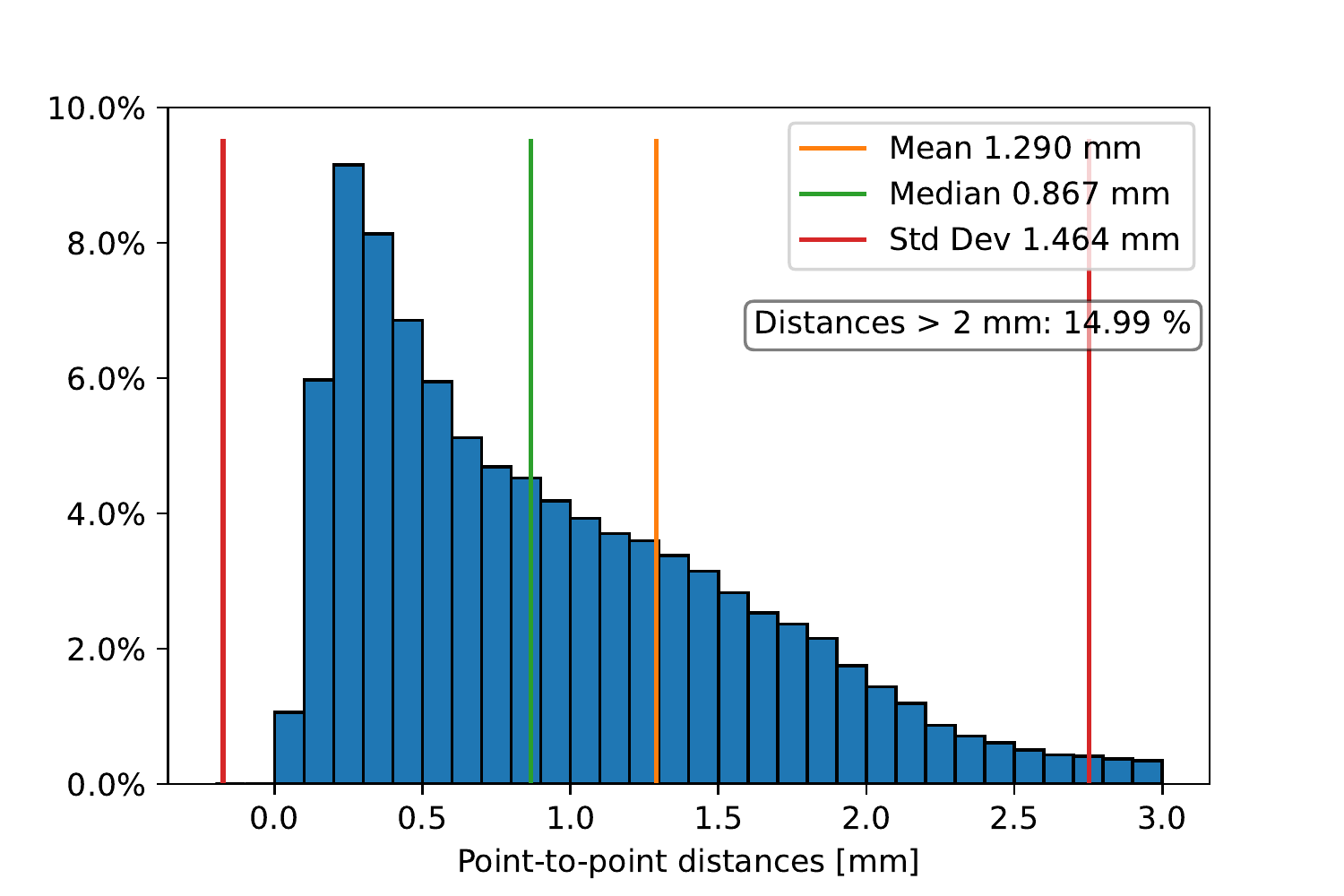}
\end{subfigure}
\caption{Distribution of point-to-point distances [\SI{}{mm}] from CT point cloud to the nearest neighbour in photogrammetric cloud (left: CT Cuypers vs. photogrammetric Hofmans, right: CT Hofmans vs. photogrammetric Cuypers).}
\label{fig:Appendix}
\end{figure}

\noindent As the two instruments to be compared have different sizes, we did not consider the scaling factor $K$ in the matching problem (see Equations \ref{eq:Min_D} and \ref{eq:RBT}). Indeed, it would not make sense to favour the results with an artificial transformation. However, for a fair comparison between CT and photogrammetric representations, we applied the scaling of Table \ref{tab:MinProb} so that the sizes of the photogrammetric instruments still correspond to their size in CT representation. Specifically, the photogrammetric Hofmans and Cuypers were scaled respectively by a factor $1.024$ and $1.029$. \\

\begin{figure}[H]
    \centering
    \hspace{0.4 cm}
    \begin{subfigure}{0.48\textwidth}
    \includegraphics[height = 4 cm]{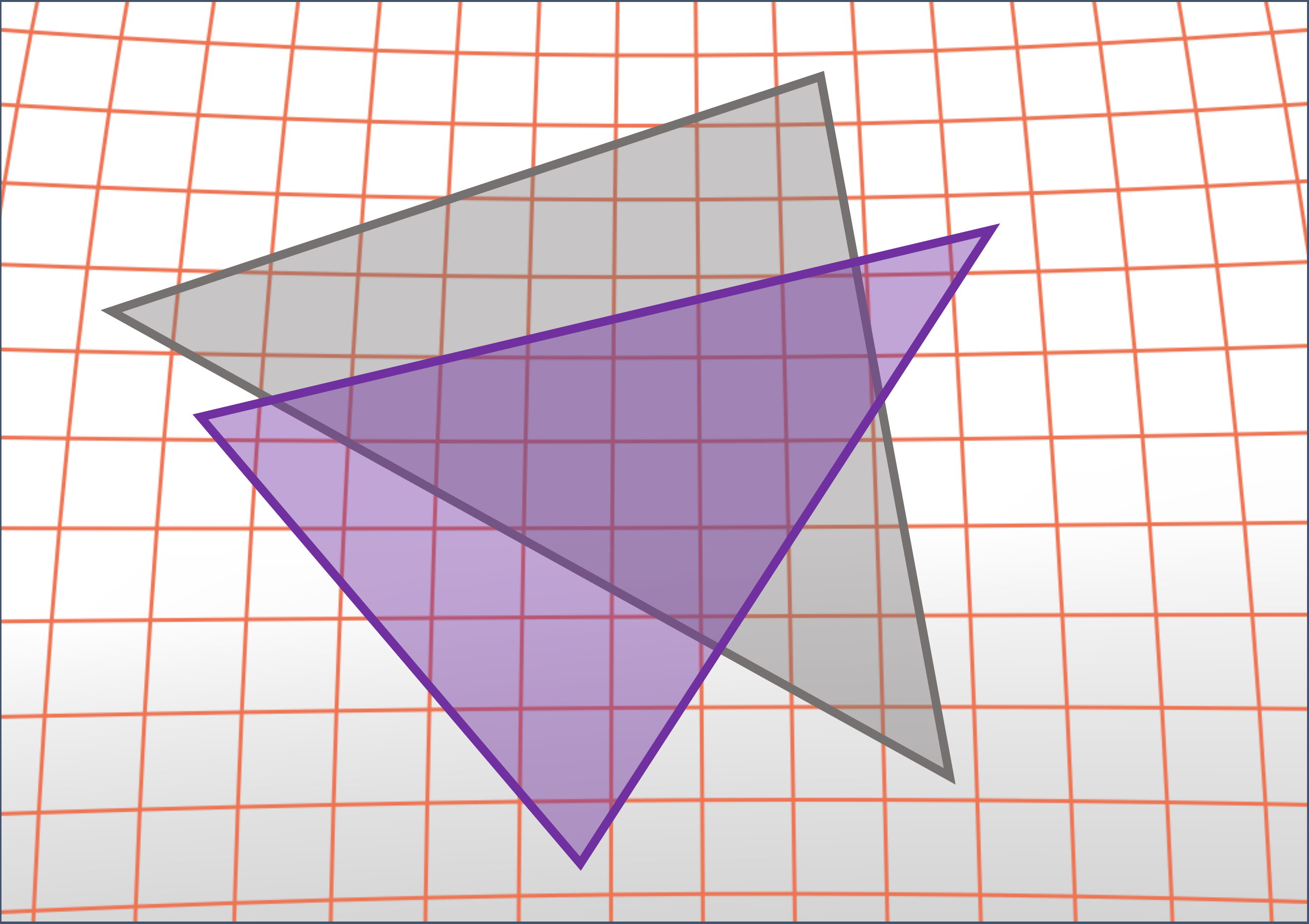}
    \end{subfigure}
    \hspace{-0.4cm}
    \begin{subfigure}{0.48\textwidth}
    \includegraphics[height = 4 cm]{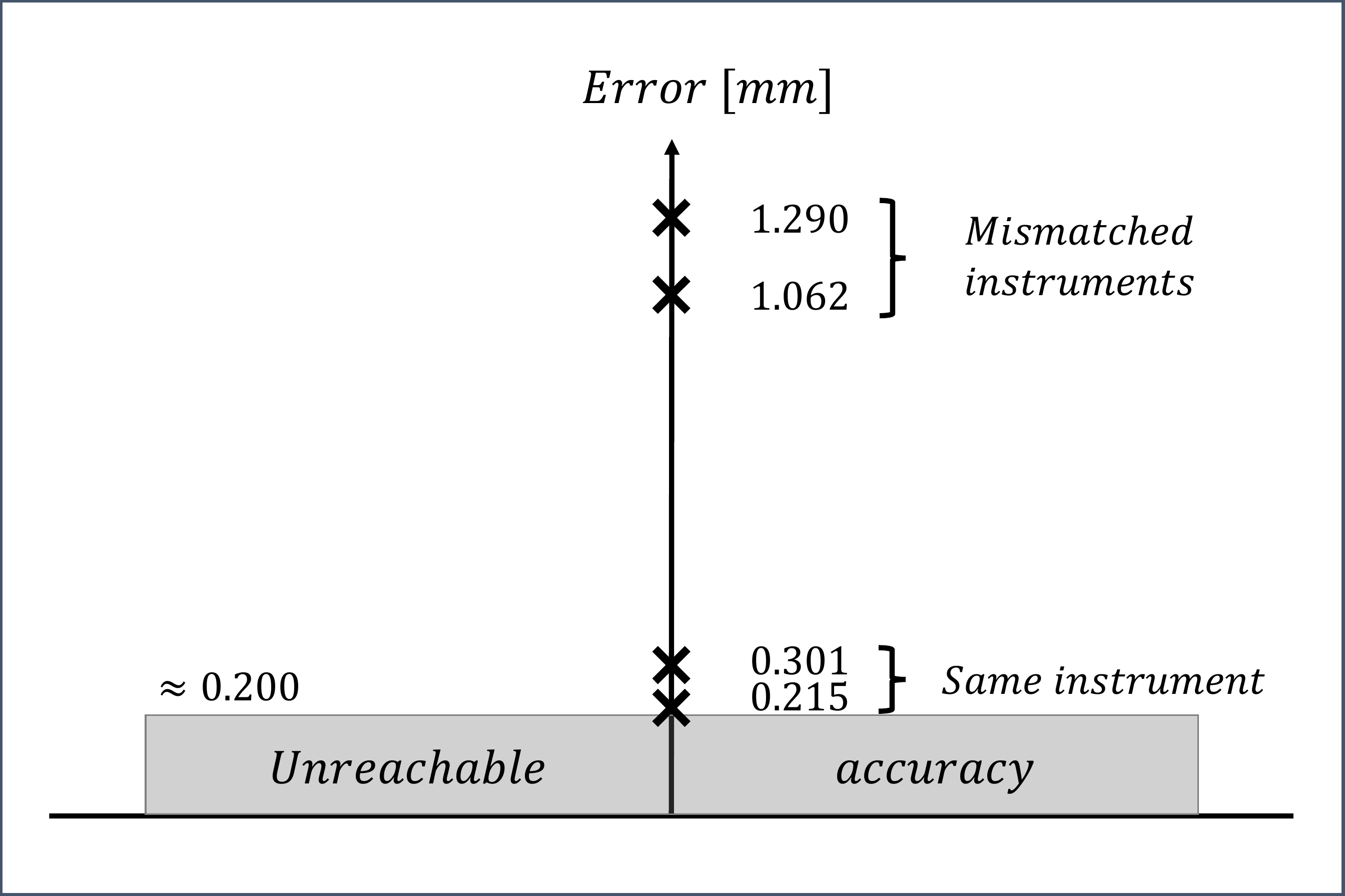}
    \end{subfigure}
    \caption{Example of two acquisition techniques that both detect points belonging to exactly the same surface but sampled at different locations (left). Representation of the four comparisons errors (CT scan vs. photogrammetry) with respect to an unreachable accuracy of about \SI{0.2}{\mm} (right).}
    \label{fig:poor_fit}
\end{figure}

\noindent Comparing two different violins does not make much sense in the study of musical instruments and should reveal a poor correspondence. However, the average point-to-point distances for the two comparisons are \SI{1.062}{\mm} and \SI{1.290}{\mm}, a surprisingly small value despite the fact that the two instruments are different and one of them has been reduced. To explain this, we first observe in Figure \ref{fig:Appendix} that the heat maps and histograms clearly indicate a much poorer match than in Figure \ref{fig:Data}. In addition, we realised that even if the photogrammetric mesh was perfectly describing the sound board, its vertices cannot be expected to be located exactly in the same places as the vertices of the CT mesh, as illustrated in Figure \ref{fig:poor_fit} (left). The average length of the edges in our CT meshes are equal to \SI{0.59}{mm} (Hofmans) and \SI{0.51}{mm} (Cuypers). Assuming that the average distance between two independent meshes of the same object can not be significantly smaller than the third of that average edge length ($\approx$ \SI{0.20}{mm}, see right of Figure \ref{fig:poor_fit}), we conclude that the match of both the Hofmans and the Cuypers violins was excellent in Section \ref{Error assessment and validation}, and that the average error between mismatched instruments shown in this appendix is actually quite significant.

\bibliographystyle{elsarticle-num} \bibliography{ISPRS.bib}




\end{document}